\documentclass[10pt,twocolumn,letterpaper]{article}

\usepackage[pagenumbers]{cvpr} 

\usepackage[accsupp]{axessibility}

\usepackage{graphicx}
\usepackage{amsmath}
\usepackage{amssymb}
\usepackage{booktabs}
\usepackage{multirow}
\usepackage{algorithm, algorithmic}
\usepackage{soul}

\usepackage[pagebackref,breaklinks,colorlinks]{hyperref}

\usepackage[capitalize]{cleveref}
\crefname{section}{Sec.}{Secs.}
\Crefname{section}{Section}{Sections}
\Crefname{table}{Table}{Tables}
\crefname{table}{Tab.}{Tabs.}

\def\cvprPaperID{11306} 
\def\confName{CVPR}
\def\confYear{2023}

\begin{document}

\title{NVTC: Nonlinear Vector Transform Coding}

\author{
Runsen Feng ~
Zongyu Guo ~
Weiping Li ~
Zhibo Chen \\ 
University of Science and Technology of China \\ 
{\tt\small \{fengruns, guozy\}@mail.ustc.edu.cn, \{wpli, chenzhibo\}@ustc.edu.cn}
}
\maketitle

\begin{abstract}

In theory, vector quantization (VQ) is always better than scalar quantization (SQ) in terms of rate-distortion (R-D) performance \cite{VQadvantage}. 
Recent state-of-the-art methods for neural image compression are mainly based on nonlinear transform coding (NTC) with uniform scalar quantization, overlooking the benefits of VQ due to its exponentially increased complexity. 
In this paper, we first investigate on some toy sources, demonstrating that even if modern neural networks considerably enhance the compression performance of SQ with nonlinear transform, there is still an insurmountable chasm between SQ and VQ.
Therefore, revolving around VQ, we propose a novel framework for neural image compression named Nonlinear Vector Transform Coding (NVTC). 
NVTC solves the critical complexity issue of VQ through (1) a multi-stage quantization strategy and (2) nonlinear vector transforms. 
In addition, we apply entropy-constrained VQ in latent space to adaptively determine the quantization boundaries for joint rate-distortion optimization, which improves the performance both theoretically and experimentally. 
Compared to previous NTC approaches, NVTC demonstrates superior rate-distortion performance, faster decoding speed, and smaller model size. 
Our code is available at \url{https://github.com/USTC-IMCL/NVTC}.

\end{abstract}

\section{Introduction}
\label{sec:intro}

\begin{figure}[th]
 \centering
 \begin{subfigure}{\linewidth}
\includegraphics[width=\linewidth, clip]{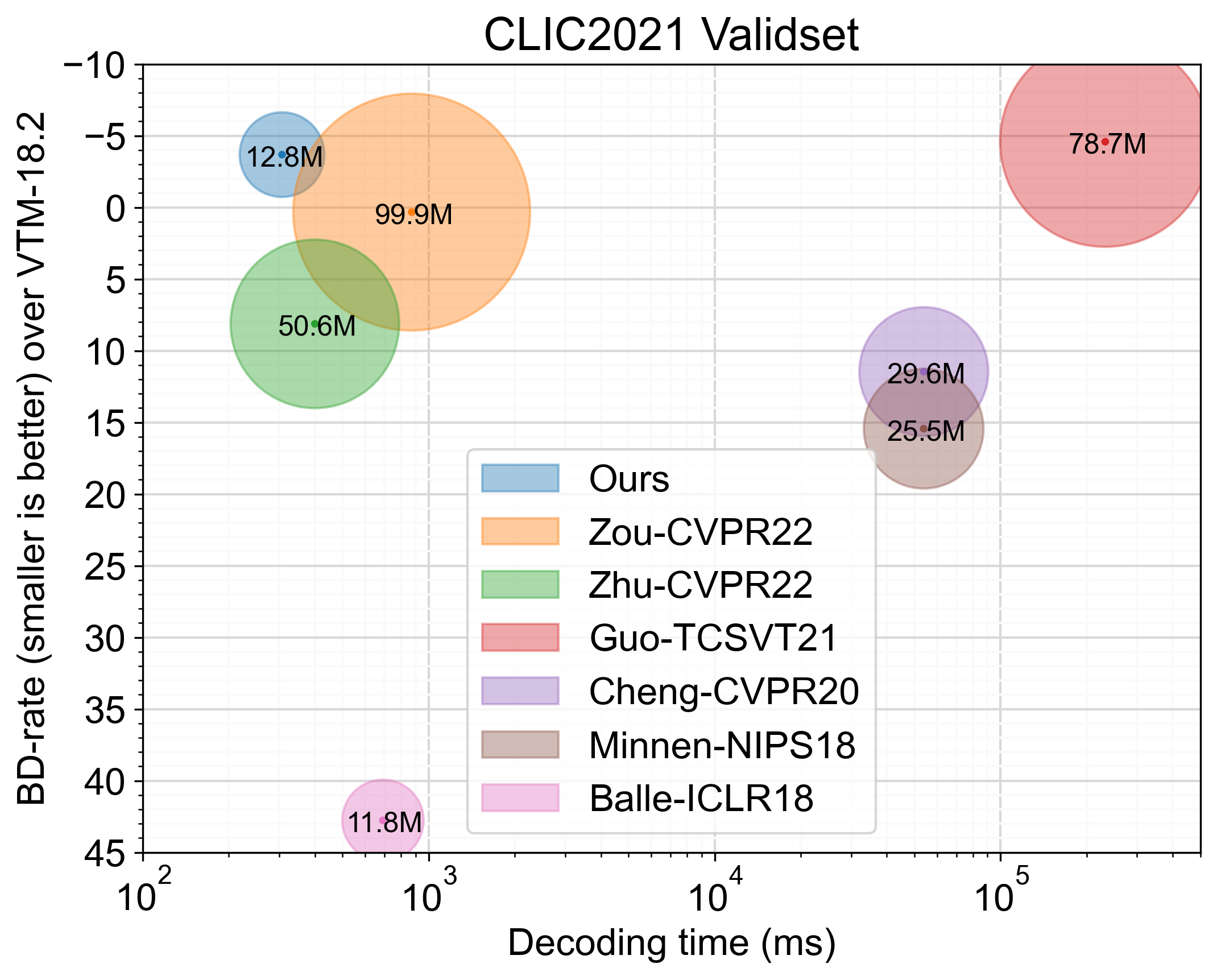}
 \end{subfigure}

 \vspace{-2mm}
 \caption{BD-rate vs. decoding time vs. model size on the CLIC2021 validation set \cite{CLIC}.
 }
\label{fig:rdc_clic_psnr}
\vspace{-2mm}
\end{figure}

\begin{figure*}[t]
 \centering
 \begin{subfigure}{\linewidth}
\includegraphics[width=\linewidth, clip, trim=0cm 0cm 0cm 0cm]{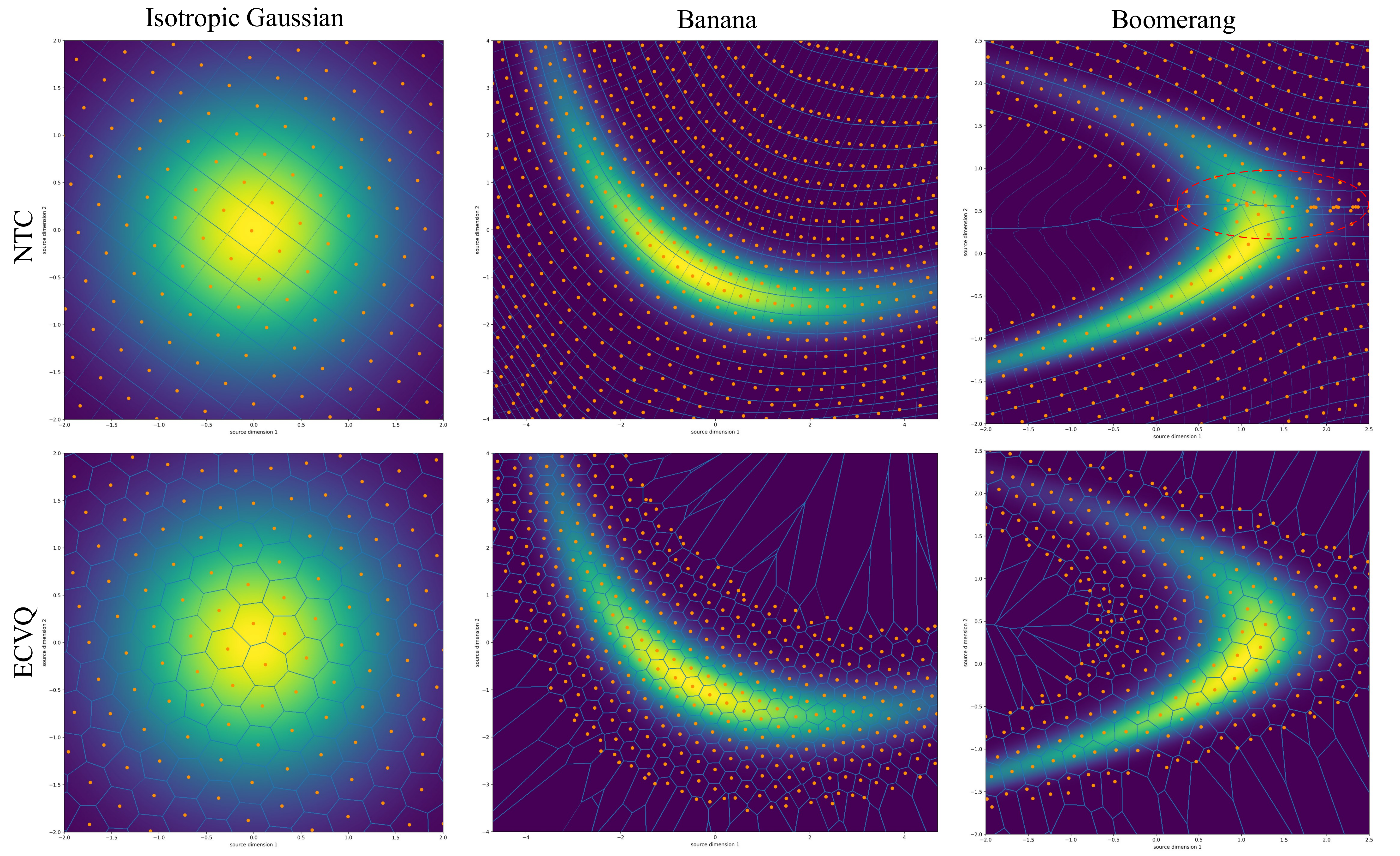}
 \end{subfigure}
 
 \caption{Quantization results for NTC (SQ with nonlinear transform) and ECVQ (entropy-constrained VQ) on 2-d distributions. Blue lines represent quantization boundaries and orange points represent quantization centers (codewords). ``Isotropic Gaussian'' refers to a 2-d  isotropic Gaussian distribution. ``Banana'' and ``Boomerang'' are two different 2-d distributions. It is observed that NTC cannot achieve the space-filling advantage (i.e. learning hexagon-like quantization cells for 2-d sources) even on an isotropic gaussian distribution. Moreover, NTC's decorrelation capability is insufficient as source correlation becomes more complex. For example, quantization boundaries collide in the red circle of "Boomerang", leading to a performance drop. The corresponding BD-PSNR results are shown in Table~\ref{tab:bdpsnr-2}.
 }
\label{fig:vis_quant}
\vspace{-3mm}
\end{figure*}

\begin{figure*}[t]
 \centering
 \begin{subfigure}{\linewidth}
\includegraphics[width=\linewidth, clip, trim=0cm 0cm 0cm 0cm]{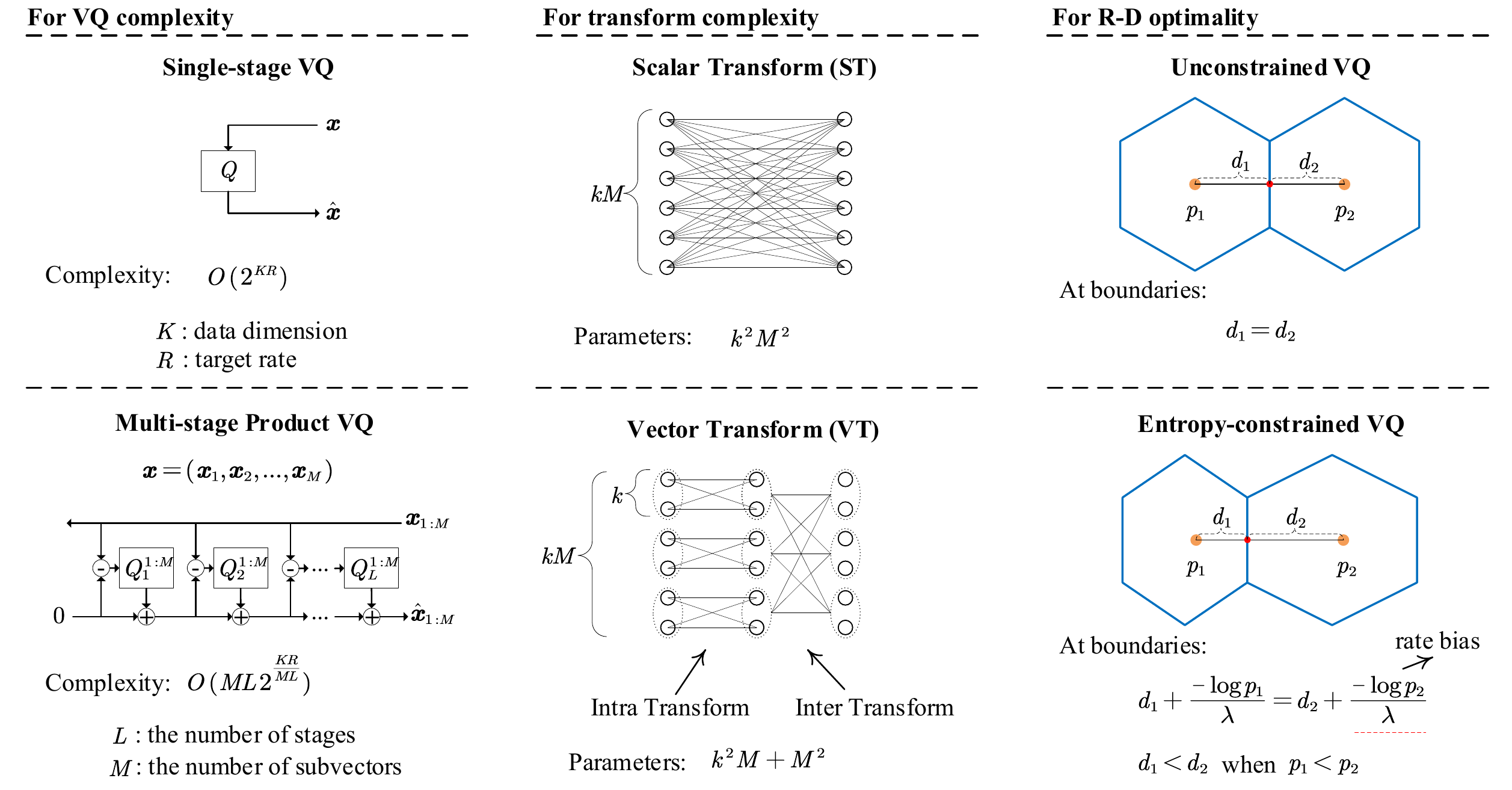}
 \end{subfigure}

 \caption{ Three key points to design a practical vector quantizer. For VQ complexity (left), we suggest a hybrid VQ structure called multi-stage product VQ which reduces the VQ complexity from $O(2^{KR})$ to $O(ML2^{\frac{KR}{ML}})$. For transform complexity (middle), we use vector transform instead of scalar transform to remove inter-vector redundancy. For RD optimality (right), we find that ECVQ~\cite{ECVQ} is essential for the joint rate-distortion optimization, which is neglected in previous works~\cite{Zhu-CVPR2022, Agustsson-NIPS2017, VTC-1995, VQVAE}}.
\label{fig:startingpoints}
\vspace{-4mm}
\end{figure*}
 
Recent works based on nonlinear transform coding (NTC)~\cite{NTC} have achieved remarkable success in neural image compression \cite{Minnen-NIPS2018,Cheng-CVPR2020}.
Unlike these traditional image codecs that employ linear transform such as discrete cosine transform (DCT), NTC is constructed with the nonlinear transform layers and optimized with data-driven techniques, where the modern neural networks present excellent capability in both encoding/decoding transform and entropy estimation \cite{Cheng-CVPR2020,Guo-TCSVT2021,Minnen-ICIP2020}.  
Most NTC methods apply scalar quantization (SQ) to discretize the latent variables and use the additive uniform noise to approximate the quantization error during training \cite{balle2016end}. However, in the era of 1990s, it has already been known that vector quantization, in spite of it exponentially increased complexity, is always better than SQ in terms of rate-distortion (RD) performance \cite{VQadvantage}. It
inspires us to design a novel neural image compression model to fully leverages vector quantization.

Vector quantization (VQ)~\cite{VQ} is designed to map a continuous source distribution to a set of discrete vectors. The discrete nature of VQ has been successfully applied in generative models to avoid the “posterior collapse” issue, including these well-known image synthesis models such as VQVAE~\cite{VQVAE}, VQGAN~\cite{VQGAN} and text-to-image synthesis models such as DALLE\cite{DALLE}, latent diffusion \cite{rombach2022high}. However, 
if we go back to the basic requirement of quantization, we will find that VQ offers unique advantages in terms of rate-distortion performance, particularly the space-filling advantage and the memory advantage \cite{VQadvantage}.

Given a source distribution, the goal of quantization (no matter SQ or VQ) is to determine the quantization centers and boundaries, and then assign indices to denote these separated quantization regions/cells. Combining these regions fills the whole space of source distribution. The space-filling advantage of VQ against SQ is related to the sphere packing problem in geometry \cite{SP,SP8,SP24}. 
If we compare the quantization results of SQ and VQ, as shown in Figure~\ref{fig:vis_quant}, we will find that even for a simple isotropic Gaussian distribution, SQ with nonlinear transform cannot learn to approximate hexagon-like quantization cells, where the hexagon is the polytope with the best space-filling properties in 2-d space. 
Under the high-rate assumption, the gain of space-filling advantage is about 1.53 dB as the dimension approaches infinity~\cite{VQadvantage,conway1982voronoi}. Following this conclusion, we experimentally provide the BD-PSNR results by comparing SQ with nonlinear transform to VQ on isotropic Gaussian distributions in Table~\ref{tab:bdpsnr-1}. 
In addition, to reduce the redundancies of data distributions, existing NTC methods (SQ with nonlinear transform) rely on highly expensive nonlinear transform~\cite{Zhu-ICLR2022, Zou-CVPR2022, Cheng-CVPR2020} and context-based auto-regressive entropy models~\cite{Guo-TCSVT2021, Minnen-ICIP2020}. 
However, different from NTC methods, VQ has superior decorrelation ability, which is known as the memory advantage of vector quantizers. 
This advantage is more obvious when quantizing complex source distributions, such as the Boomerang distribution in Figure~\ref{fig:vis_quant} (especially in the red circle area). 

In this paper, we build a novel framework that applies modern neural networks to leverage the space-filling advantages and memory advantages of VQ for image compression. 
We propose nonlinear vector transform coding (NVTC), which achieves encouraging rate-distortion performance with relatively low coding complexity. 
Specifically, as shown in Figure~\ref{fig:startingpoints}, we introduce three key points to design a practical VQ, including 
1) a multi-stage product VQ rather than a single-stage VQ to reduce the exponentially increased complexity, 
2) nonlinear vector transform rather than scalar transform to remove redundancy between sub-vectors with fewer parameters, and 
3) entropy-constrained VQ rather than unconstrained VQ to achieve superior R-D optimality and joint optimization of latent-space VQ models.

For \textbf{the first point},
many well-known VQ variants have been proposed in recent decades, such as product VQ~\cite{ProductVQ, PQ}, multi-stage VQ~\cite{MSVQ}, tree-structured VQ~\cite{TSVQ} and lattice VQ~\cite{LVQ}. 
Although tree-structured and lattice VQ offer fast encoding speeds, they do not reduce the storage complexity of codebooks or entropy-coding frequency tables
In this paper, we suggest a hybrid VQ structure that incorporates both product VQ and multi-stage VQ, as shown in the left column of Figure~\ref{fig:startingpoints}. The quantization procedure comprises multiple stages, and each stage employs multiple independent low-dimension quantizers to compress the subvectors of the input vector.
As the number of stages and subvectors increases, the proposed multi-stage product VQ exhibits a significant decrease in complexity.

While the intra-vector redundancy (i.e. the redundancy inside each subvector) can be removed by vector quantization, the inter-vector redundancy (i.e. the redundancy between subvectors) is still overlooked. 
Therefore, \textbf{our second point} focuses on efficiently eliminating inter-vector redundancy.
Transform VQ~\cite{STVQ} introduces a linear transform for decorrelation and performs product quantization on the transformed coefficients. 
Similar coding structures are observed in recent learning-based VQ methods~\cite{Agustsson-NIPS2017, Zhu-CVPR2022}, which are improved by learnable nonlinear transform with superior decorrelation capabilities.  
However, the transform used in these works is designed to decorrelate scalar components, which is computationally inefficient for vector decorrelation. 
The intra-vector redundancy, which is intended to be removed by VQ, might be partially reduced in advance by the scalar transform.
Therefore, certain parts of the scalar transform could be eliminated to improve computational efficiency.
Motivated by the linear vector transform~\cite{VTC-1991, VTC-1993, VTC-1995}, we propose a new VT variant that decouples a fully-connected scalar transform into two light-weight parts: intra-transform and inter-transform.
In the middle of Figure~\ref{fig:startingpoints}, we provide a simple comparison between the scalar transform and the proposed vector transform. 
We further stack the single-layer VT to build a powerful nonlinear vector transform. The differences between our VT and the linear VT are discussed in Section~\ref{sec:VT}.

Regarding \textbf{the third point}, we emphasize that the quantization process (either SQ or VQ) used in most previous methods~\cite{NTC, Zhu-CVPR2022, Agustsson-NIPS2017, VTC-1995, VQVAE} (including VQVAE) is not entropy-constrained, which is theoretically suboptimal for rate-distortion performance. 
In the right of Figure~\ref{fig:startingpoints}, we provide a quantization illustration of unconstrained VQ and entropy-constrained VQ (ECVQ~\cite{ECVQ}), where unconstrained VQ determines the quantization boundaries (blue line) using the nearest neighbor search. 
ECVQ introduces an additional rate bias $\frac{-\log{p_i}}{\lambda}$ in the quantization process, which shifts the quantization boundaries from the high-probability region to the low-probability region. In other words, ECVQ search the codewords with the best RD performance, instead of just the neighboring codewords.
ECVQ provides an optimal VQ encoding process described in Section~\ref{sec:related_works-ECVQ}.
With the help of latent-space ECVQ, we design a training strategy for joint RD optimization.
Instead of manually controlling the RD trade-off by varying codebook size~\cite{Zhu-CVPR2022}, our model can learn layer-adaptive bit allocation. 

Our contributions can be summarized as 1) investigating on VQ advantages over SQ with nonlinear transform based on empirical results on some toy sources, 2) presenting a VQ-based coding scheme named nonlinear vector transform coding (NVTC) with three technical contributions that effectively leverages VQ while keeping complexity low, and 3) demonstrating that NVTC offers superior rate-distortion performance, faster decoding speed and smaller model size, compared with previous neural image codecs.

\section{Related Works \& Background} \label{sec:related_works}
The task of lossy coding aims to map the k-dimensional Euclidean space $\mathbb{R}^k$ of a data source into a set of vectors $\boldsymbol{c}=\{\boldsymbol{c}_i \in \mathbb{R}^k| i=0, 1, ..., N-1\}$. At the encoder side, each sample vector ${\boldsymbol{x}}$ can be represented by a reproduction vector $\boldsymbol{\hat{x}}=\boldsymbol{c}_i\in \boldsymbol{c}$. The vector index $i$ is losslessly coded by an entropy model $P(i)$, which is an estimate of the probability distribution of the index. Both the vector set $\boldsymbol{c}$ and entropy model $P$ are known to the decoder. Thus, $\boldsymbol{\hat{x}}$ can be reconstructed by the entropy decoded index $i$ at the decoder side. The average code length $R$, also called rate, is given by the cross entropy between the actual distribution and the estimated distribution: $R=\mathbb{E}_{\boldsymbol{x}}{\left[-\log{P(i)}\right]}$. And the average distortion is $\mathbb{E}_{\boldsymbol{x}} {\left[ d(\boldsymbol{x}, \boldsymbol{\hat{x}})\right]}$ given a distortion metric $d$.

\textbf{Unconstrained VQ}.
A vector quantizer~\cite{VQ} $Q$ of dimension $k$ and size $N$ is defined as a mapping from $\mathbb{R}^k$ into a finite set $\boldsymbol{c}=\{\boldsymbol{c}_i \in \mathbb{R}^k| i=0, 1, ..., N-1\}$. 
The set $\boldsymbol{c}$ here is called codebook and the vector $\boldsymbol{c}_i$ is called codeword. 
Given a input vector ${\boldsymbol{x}}$, the encoder $\mathcal{E}$ searches for the best  codeword $\boldsymbol{c}_i$ and output the corresponding index $i = \mathcal{E}(\boldsymbol{x}) = \mathop{\arg\min}_i d(\boldsymbol{x}, \boldsymbol{c}_i)$. 
With the input index  $i$, the decoder $\mathcal{D}$ looks up the codebook to produce the reproduction vector $\boldsymbol{\hat{x}}=\mathcal{D}(i)=\boldsymbol{c}_i$. 

\textbf{ECVQ}.\label{sec:related_works-ECVQ}
In unconstrained VQ, the input $\boldsymbol{x}$ is assigned to the codeword $\boldsymbol{c}_i$ with the lowest distortion $d(\boldsymbol{x}, \boldsymbol{c}_i)$. 
However, it is suboptimal for RD loss: $\mathbb{E}_{\boldsymbol{x}}{\left[-\log{P(i)} + \lambda d(\boldsymbol{x}, \mathcal{D}(\mathcal{E}(\boldsymbol{x}))) \right]}$, where $\lambda$ is the Lagrange multiplier that controls RD trade-off. 
Entropy-constrained vector quantization (ECVQ)~\cite{ECVQ} introduces this Lagrangian formulation into the encoder function: $\mathcal{E}(\boldsymbol{x}) = \arg\min_i \left[ -\log{P(i)} + \lambda d(\boldsymbol{x}, \boldsymbol{c}_i) \right]$,  which is exactly the RD loss of a sample. 
It means that ECVQ always produces optimal quantization results for each sample, given the codebook and entropy model.
In the middle of Figure~\ref{fig:startingpoints}, we illustrate the differences between unconstrained VQ and ECVQ, where ECVQ shifts the quantization boundaries (blue line) from the high-probability region to the low-probability region.

\textbf{NTC}.
Nonlinear transform coding (NTC)~\cite{NTC} is  a nonlinear extension of transform coding~\cite{TC}, which simplifies vector quantization as scalar quantization and decorrelates data sources using transform and entropy model. NTC is the most popular coding structure in recent methods for learned image compression.
The state-of-art NTC methods improve the components of transform coding by designing powerful nonlinear transforms~\cite{Cheng-CVPR2020, Zhu-ICLR2022, Zou-CVPR2022}, complex forward-adaptive or backward-adaptive entropy models~\cite{Balle-ICLR2018, Minnen-NIPS2018, Minnen-ICIP2020, Guo-TCSVT2021}, and differentiable scalar quantization~\cite{Agustsson-NIPS2020, Yang-NIPS2020}. In our paper, we will use NTC to denote SQ with nonlinear transform.

\textbf{Latent-space VQ}.
Prior works such as transform VQ~\cite{STVQ} and vector transform coding~\cite{VTC-1995, VTC-1993, VTC-1991} (VTC) first propose to vector quantize the transform coefficients decades ago.
However, due to the high complexity of VQ and the weak decorrelation capabilities of linear transform, they draw less attention to the researchers compared to transform coding~\cite{TC}.
Recently, VQ is known to be effective to address the ``posterior collapse'' issue in many generative models~\cite{VQVAE, VQGAN, DALLE}. In these works, VQ is integrated into an autoencoder network for quantizing the bottleneck latents. 
Latent-space VQ is also considered in learned image compression~\cite{Agustsson-NIPS2017, Zhu-CVPR2022}. 
\cite{Agustsson-NIPS2017} propose a soft relation of VQ for end-to-end optimization, which is gradually annealed to real VQ during training. 
\cite{Zhu-CVPR2022} propose to replace the univariate prior with vectorized prior in existing NTC methods, where a probabilistic vector quantizer is proposed to estimate the prior parameters. 

\begin{table}[th]

\begin{center}
\begin{tabular}{ccccc}
& \bf 2d & \bf 4d & \bf 8d  & \bf 16d  \\
\hline \\
BD-PSNR (dB) & -0.15 & -0.31 & -0.47 & -0.71
\end{tabular}
\end{center}
\vspace{-2mm}
\caption{BD-PSNR (smaller is worse) of NTC over ECVQ on the isotropic Gaussian distributions with different dimensions.}
\label{tab:bdpsnr-1}
\vspace{-2mm}
\end{table}

\begin{table}[th]

\begin{center}
\begin{tabular}{cccc}
& \bf IG & \bf Banana & \bf Boomerang  \\
\hline \\
BD-PSNR (dB)  & -0.15 & -0.21 & -0.58
\end{tabular}
\end{center}

\vspace{-2mm}
\caption{BD-PSNR (smaller is worse) of NTC over ECVQ on the 2-d distributions in Figure~\ref{fig:vis_quant}. IG means Isotropic Gaussian.}
\label{tab:bdpsnr-2}
\vspace{-2mm}
\end{table}

\begin{figure*}[t]
 \centering
  \begin{subfigure}{0.51\linewidth}
 \includegraphics[width=\linewidth, trim=0cm 0cm 0cm 0cm]{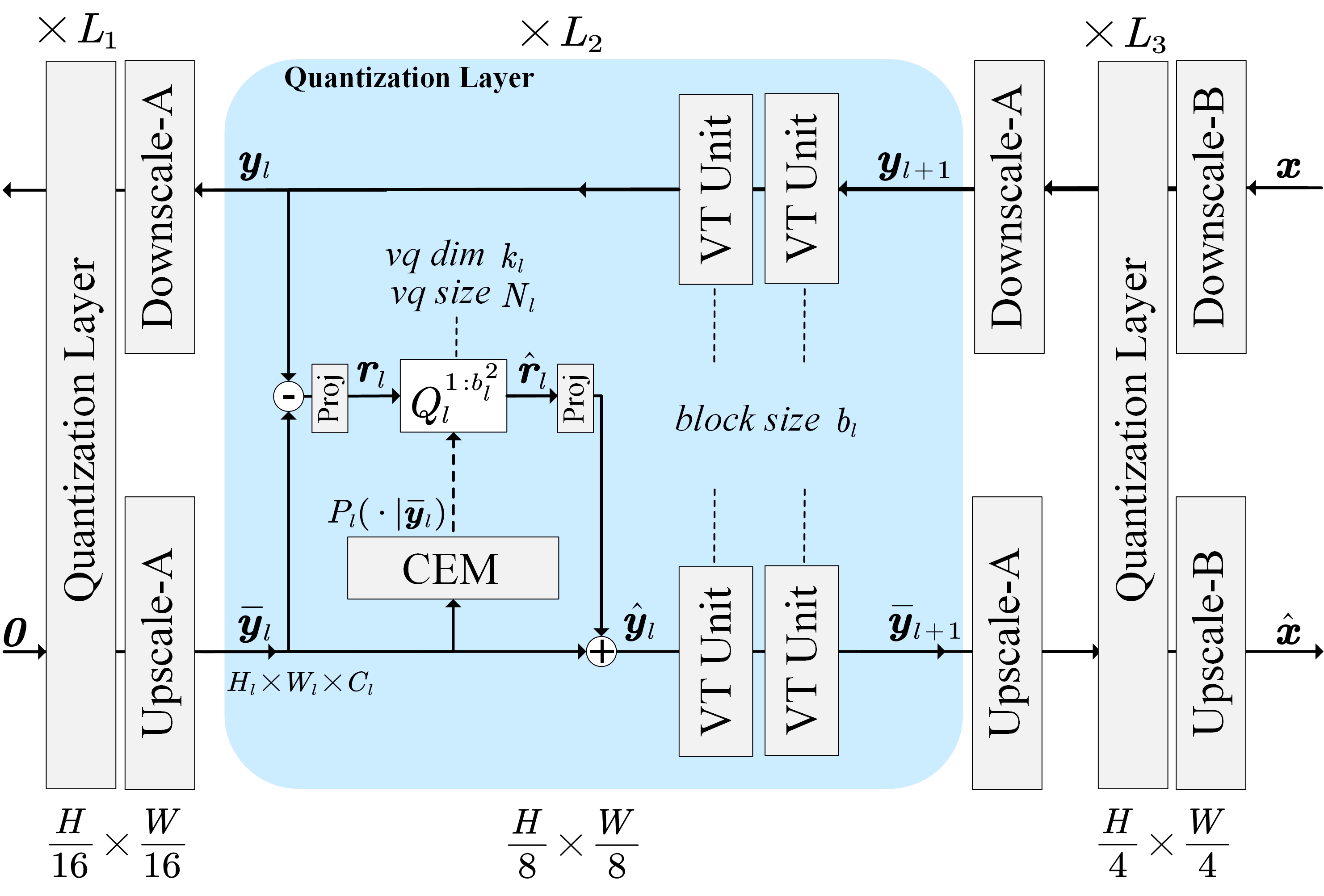}
 \label{fig:overview-a}
 \end{subfigure}
\hspace{5mm}
 \begin{subfigure}{0.44\linewidth}
 \includegraphics[width=\linewidth, trim=0cm 0cm 0cm 0cm]{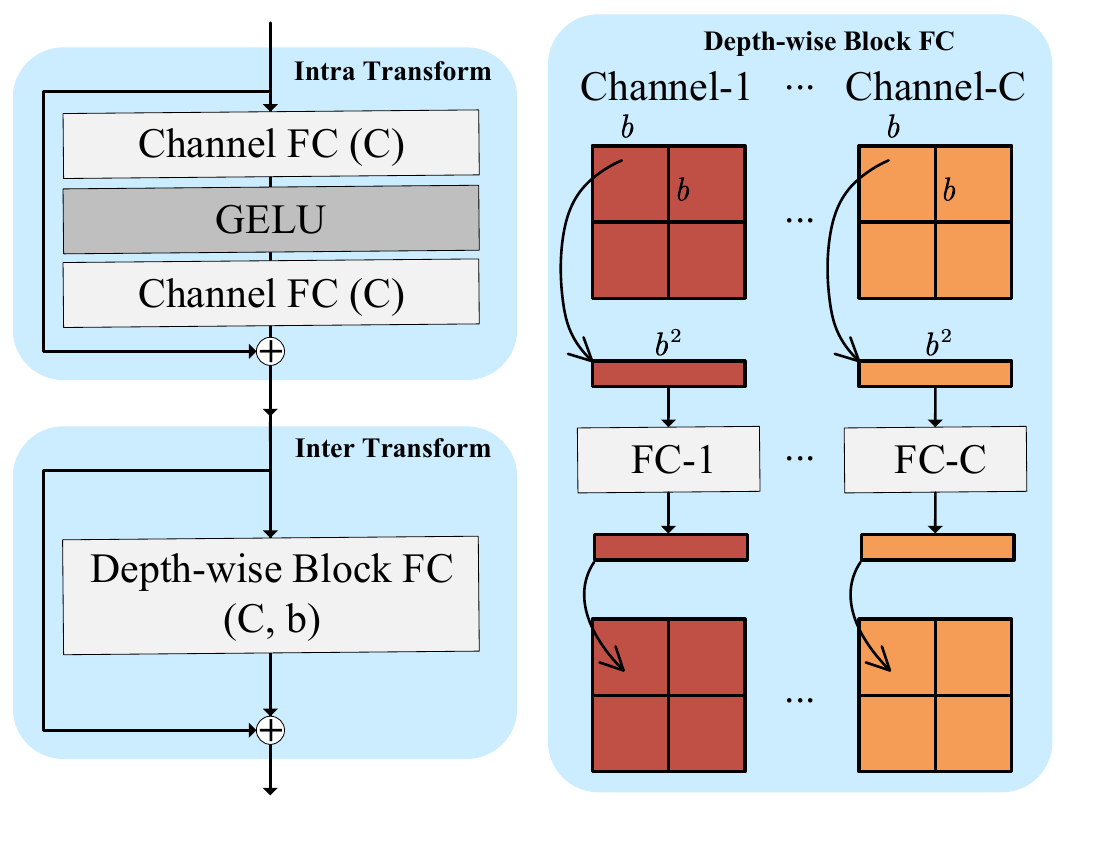}
 \label{fig:overview-b}
 \end{subfigure}
 
\vspace{-4mm}
\caption{(Left) Overview of our proposed model for image compression. VT means vector transform and CEM means conditional entropy model. Proj is a linear projection layer. CEM is described in Section~\ref{sec:CEM}. (Right) VT unit with depth-wise block FC. FC refers to a fully-connected layer, and Channel FC is a fully-connected layer along the channel axis. See Section~\ref{sec:VT} for more details.}
\label{fig:overview}
\vspace{-2mm}
\end{figure*}

\section{VQ Advantages on Toy Sources}\label{sec:toysources}
Under high-rate assumptions, vector quantization has three advantages over scalar quantization: space-filling advantage, shape advantage and memory advantage~\cite{VQadvantage}. The first advantage comes from the space-filling properties of different polytopes which is independent of the source distribution, making it a fundamental advantage of VQ over scalar quantization. The second advantage depends solely on the shape of the marginal density function, regardless of the correlation of vector components. The third one reveals the decorrelation capability, which is the major advantage when dealing with highly correlated source data. 
The shape advantage could be obtained by accurate entropy modeling.
Therefore, in this section, we focus on investigating whether NTC effectively achieves the space-filling and memory advantages. More results can be found in the supplementary.

To verify the space-filling advantage, we first conduct experiments on isotropic Gaussian distributions, where the memory advantage of VQ does not exist. 
On the left of Figure~\ref{fig:vis_quant} we visualize the quantization results on a 2d isotropic Gaussian. 
It can be observed that NTC has rectangular-like quantization cells, different from hexagon-like cells in ECVQ. 
Hexagon has better space-filling efficiency than rectangular in 2d space. 
The gain of space-filling advantage increases as the source dimension grows higher. 
Under the high-rate assumption, the approximate gain of space-filling advantage is 1.53 dB as the dimension approaches infinity~\cite{VQadvantage,conway1982voronoi}. 
We also conduct experiments to verify the empirical rate-distortion performance on 4d, 8d and 16d isotropic Gaussian. 
As shown in Table~\ref{tab:bdpsnr-1}, the BD-PSNR~\cite{BD-metric} gain of ECVQ is surprisingly 0.71 dB on 16d isotropic Gaussian, which is about 50\% of the estimated gain (i.e. 1.53 dB) on infinite-dimensional distributions. 
Considering that VQ complexity grows exponentially with dimension, a 16-d vector quantizer is sufficient to achieve the space-filling advantage under the current complexity constraints.

For memory advantage, we evaluate NTC and ECVQ on several 2d correlated sources. In Figure~\ref{fig:vis_quant} we show the results on two distributions: one is banana-like distribution named ``Banana'' that comes from~\cite{NTC}; the other is named ``Boomerang'' which has a larger bending angle than ``Banana''. We find that the BD-PSNR gains of ECVQ on these distributions are more significant than the gain of space-filling advantage alone (i.e. 0.15 dB on 2d isotropic Gaussian). Even if the local transformation of NTC is roughly orthogonalized~\cite{NTC}, there are some failure cases in the highly nonlinear density area (see red circle in Figure~\ref{fig:vis_quant}). In a word, the potential gain of memory advantage is also significant when the source density is complex.

\section{Nonlinear Vector Transform Coding}
The overview of the proposed method is shown in Figure~\ref{fig:overview}. Following the structure of multi-stage product VQ in Figure~\ref{fig:startingpoints}, NVTC comprises multiple quantization layers, where quantization is performed sequentially from the leftmost layer to the rightmost layer. 
Motivated by the hierarchical structure in VAE-based methods~\cite{Balle-ICLR2018, VQVAE2}, we integrate  Downscale/Upscale layers between the quantization layers to facilitate global-to-local quantization. 
In other words, NVTC first quantizes the global vector representation derived from big image patches and then quantizes the local vector representation derived from small image patches.

Let's take a deeper look at the quantization layer $l$. 
Note that $N_l$ is the VQ size, $k_l$ is the VQ dimension and $b_l$ is the block size. 
The vector representation $\boldsymbol{y}_l$ has a spatial size of $H_l \times W_l$ and a channel size of $C_l$. 
Each channel tensor is considered as a subvector. The vector transform and product vector quantizers $Q_l^{1:b_l^2}$ are shared across different spatial blocks of size $b_l \times b_l$, and for each block, $b_l^2$ sub-vectors are processed.

Given $\boldsymbol{y}_{l+1}$ from layer $l+1$, $\boldsymbol{y}_{l}$ is obtained by $\boldsymbol{y}_{l} = f_l(\boldsymbol{y}_{l+1})$, where $f_l$ is vector transform layer that consists of two vector transform units with a block size $b_l$.
With the output $\boldsymbol{\bar{y}}_{l}$ of the layer $l-1$, the residual vectors $\boldsymbol{r}_l = Lin1(\boldsymbol{y}_{l} - \boldsymbol{\bar{y}}_{l})$ are compressed by an entropy-constrained product vector quantizer $Q_l^{1:b_l^2}$ with the conditional vector prior $P_l(\cdot|\boldsymbol{\bar{y}}_{l})$. $Lin1$ is a linear projection layer that changes the tensor shape from $H_l \times W_l \times C_l$ to $H_l \times W_l \times k_l$. 
Given the dequantized residual $\boldsymbol{\hat{r}}_l$, the reconstructed vector representations are obtained by $\boldsymbol{\hat{y}}_{l}=Lin2(\boldsymbol{\hat{r}}_l) + \boldsymbol{\bar{y}}_{l}$ with a shape of $H_l \times W_l \times C_l$, where $Lin2$ is a linear projection layer. 
Then $\boldsymbol{\hat{y}}_{l}$ is transformed to  $\boldsymbol{\bar{y}}_{l+1}=f_l(\boldsymbol{\hat{y}}_{l})$ and fed into the next quantization layer $l+1$.

\subsection{Nonlinear Vector Transform}\label{sec:VT}
Vector transform (VT) is not simple nonlinear transform. Although VQ is always superior to scalar quantization, the high complexity hinders its practical application
(such as the huge number of codewords). Therefore, the
method of VT is designed to cooperate with VQ, which ensures the complexity is acceptable. In our initial experiments, we found that directly using the linear vector transform~\cite{VTC-1995, VTC-1993, OnVT} led to spatial aliasing and a significant performance drop. We identified two primary reasons: 1) linear transform (e.g., block-DCT) cannot handle nonlinear inter-vector correlations, and 2) intra-vector transform is crucial for obtaining effective vector representations. Therefore, we propose a novel vector transform unit comprising two components: intra transform for nonlinear representation learning and inter transform for inter-vector decorrelation.
As shown in the right of Figure~\ref{fig:overview}, the intra transform consists of two fully-connected (FC) layers along the channel axis and one activation function. The inter transform contains a Depth-wise Block FC layer that can be regarded as a block-DCT combined with channel-wise learnable transform matrices.

\subsection{Conditional Entropy Model}\label{sec:CEM}
For a single-stage VQ of dimension $k$ with codebook size $N$, the entropy model $P$ of the codeword index $i$ is a discrete probability distribution $P(i) = p_i$, where $p_i>0$ and $\sum_{i=1}^N p_i=1$. 
Following previous work~\cite{Agustsson-NIPS2017, NTC}, we parameterize $p_i$ with the Softmax function and unnormalized logits $\boldsymbol{w}=(w_1, w_2, ..., w_N)$: 
\begin{equation}
\begin{aligned}
p_i = \frac{e^{-w_i}}{\sum_{j=1}^N e^{-w_j}}.
\end{aligned}
\label{eq:ECVQ-dec}
\end{equation}
For a multi-stage product VQ like NVTC, we propose a conditional entropy model (CEM) to further remove the inter-vector redundancy.
Specifically, at layer $l$, CEM estimates the spatial-adaptive logits $\boldsymbol{w}_l \in \mathbb{R}^{H_l \times W_l \times N_l} $ by using the output $\boldsymbol{\bar{y}}_l \in \mathbb{R}^{H_l \times W_l \times C_l} $ of layer $l-1$ as:
\begin{equation}
\begin{aligned}
\boldsymbol{w}_l = \Phi(\boldsymbol{\bar{y}}_l),
\end{aligned}
\label{eq:ECVQ-dec}
\end{equation}
where $\Phi$ is a neural network described in the supplementary material. Then the conditional probability $P_l(\cdot|\boldsymbol{\bar{y}}_l)$ is calculated by the Softmax function.

\subsection{RD optimization with Latent-Space ECVQ}\label{sec:RDOwECVQ}
Due to the gradient of VQ being almost zero everywhere, it is hard to end-to-end optimize both vector quantization and nonlinear transform for the rate-distortion trade-off. 

Some recent works for generative modeling like VQVAE~\cite{VQVAE, VQGAN} use the straight-through estimator (STE) to pass gradients from the decoder transform to the encoder transform, and introduce additional losses to optimize the codebook. 
Recent works~\cite{Agustsson-NIPS2017, Zhu-CVPR2022} for image compression utilize Softmax or Gumbel Softmax to approximate the VQ gradients.
However, the joint RD trade-off is rarely considered in these works. 
For example, the rate-distortion trade-off in~\cite{Zhu-CVPR2022} is controlled by varying codebook size, where a larger codebook size corresponds to a higher bitrate. 
In this paper, similar to VQVAE~\cite{VQVAE}, we also use STE to pass gradients through VQ. 
Then we propose a new rate-distortion loss to optimize the methods with latent-space ECVQ. The single-layer loss is as follows:

\begin{equation}
\begin{aligned}
\mathcal{L} =  
& \lambda \mathrm{E}_{\boldsymbol{x}} {\left[ d(\boldsymbol{x}, \boldsymbol{\hat{x}})\right]} + \mathrm{E}_{\boldsymbol{x}} {\left[-\log{P(I)}\right]} \\
& + \beta \mathrm{E}_{\boldsymbol{x}} {\left[ d_1(\boldsymbol{y}, \boldsymbol{\hat{y}})\right]}
\end{aligned}
\label{eq:loss-vq}
\end{equation}
where $\boldsymbol{y}$ is the latent to be quantized,  $I$ is the codeword index from ECVQ, $d$ is the pixel-space distortion metric, and $d_1$ is the latent-space VQ distortion metric. 
$\lambda$ controls the rate-distortion trade-off. $\beta$ controls the trade-off between $d$ and $d_1$. 
The encoding process of ECVQ has been described in Section~\ref{sec:related_works-ECVQ}.
For multi-layer NVTC, the loss is as follows:

\begin{equation}
\begin{aligned}
\mathcal{L} =  
&\lambda \mathrm{E}_{\boldsymbol{x}} {\left[ d(\boldsymbol{x}, \boldsymbol{\hat{x}})\right]} + \sum_l \mathrm{E}_{\boldsymbol{x}} {\left[-\log{P_l(I_l | \boldsymbol{\bar{y}}_l)}\right]}  \\
&+ \beta \sum_l \mathrm{E}_{\boldsymbol{x}} {\left[ d_1(\boldsymbol{r}_l, \boldsymbol{\hat{r}}_l)\right]}
\end{aligned}
\label{eq:loss-msvq}
\end{equation}
\begin{algorithm}[h] 

\caption{Progressive Initialization of NVTC}
\label{alg:MSVQinit}                      
\begin{algorithmic}[1]              
\REQUIRE Input $\boldsymbol{x}$; The vector quantizers $\{Q_l\}_{l=1}^L$; Encoder transform $f_{1:L}$; Decoder Transform $g_{1:L}$; Iteration milestones $T_{1:L}$, ($T_i\leq T_j$ when $i < j$).

\FOR{$t \leftarrow 1$ to $T_{init}$}
\STATE $\boldsymbol{\bar{y}}_{1} \leftarrow \boldsymbol{0}$; transform $\boldsymbol{x}$ into  $\boldsymbol{y}_{1:L}$ using $f_{1:L}$
\FOR{$l \leftarrow 1$ to $L$}
\IF{$t > T_l$}
\STATE $\boldsymbol{\hat{y}}_l \leftarrow Q_l(\boldsymbol{y}_l - \boldsymbol{\bar{y}}_{l}) + \boldsymbol{\bar{y}}_{l} $
\ELSE
\STATE $\boldsymbol{\hat{y}}_l \leftarrow \boldsymbol{\bar{y}}_{l}$
\ENDIF
\STATE $\boldsymbol{\bar{y}}_{l+1} \leftarrow g_{l}(\boldsymbol{\hat{y}}_{l}) $
\ENDFOR
\STATE $\boldsymbol{\hat{x}} \leftarrow \boldsymbol{\bar{y}}_{L+1}$; update model using Equation~\ref{eq:loss-msvq}.
\IF{$t \mod \delta{T} =0$}
\STATE reinitialize low-frequency codewords.
\ENDIF
\ENDFOR 
\end{algorithmic}
\end{algorithm}

\begin{figure*}[t]
 \centering
 \begin{subfigure}{\linewidth}
\includegraphics[width=\linewidth, clip]{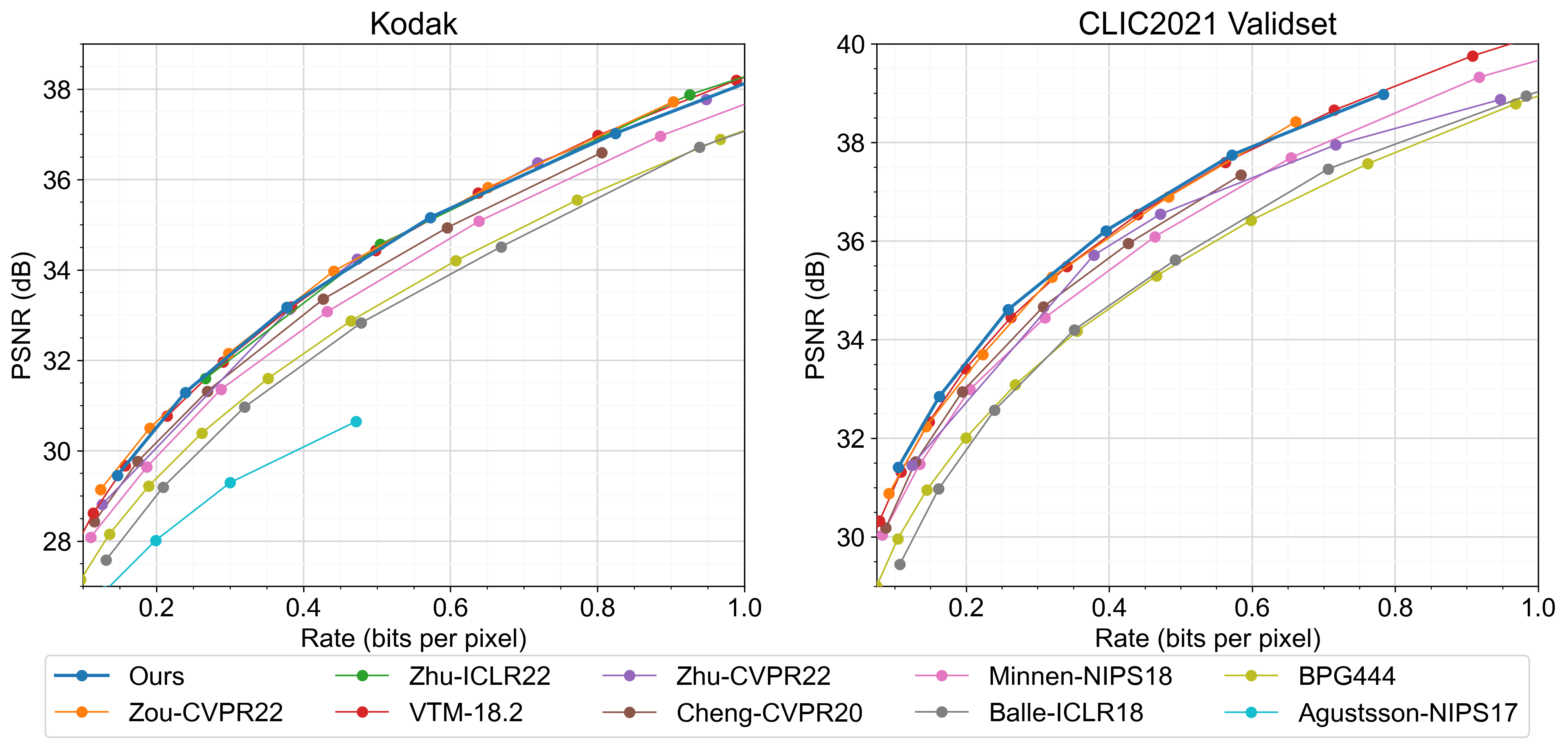}
 \end{subfigure}
 
 \vspace{-3mm}
 \caption{Rate-distortion results on kodak (left) and CLIC2021 valid set (right). The proposed method achieves state-of-the-art compression performance.}
\label{fig:rd_kodak_clic_psnr}
\vspace{-3mm}

\end{figure*}

In practice, the initialization of VQ codebooks has a significant impact on training efficiency~\cite{kmeans++, VQinit}.
We find this issue becomes more serious in the training of multi-stage VQ.
Therefore, instead of random initialization, we propose a progressive initialization method shown in Algorithm~\ref{alg:MSVQinit}. For simplicity, we omit the notation of product VQ, linear projection and Downscale/Upscale layers.
During the initialization process, ECVQ is replaced with VQ, and the conditional entropy model is removed. 
The vector quantizers are initialized sequentially.
To make full use of every codeword, we randomly reinitialize low-frequency codewords using high-frequency codewords for every $\delta{T}$ iterations.

\section{Experiments}\label{sec:experiments}

\subsection{Experiment Setup}
\textbf{Datasets.}
The image compression models were trained on COCO~\cite{Coco} train2017 set, which contains 118,287 images. Training Images are randomly cropped into 256$\times$ 256 patches. 
For evaluation, we use Kodak dataset~\cite{Kodak} which contains 24 images with 768 $\times$ 512 pixels. 
We also evaluate our models on CLIC2021 validset~\cite{CLIC} with 41 high-resolution images with a size of about 2000 $\times$ 1000.
Details regarding the toy sources are provided in the supplementary.

\textbf{Metric.}
We measure the quality of reconstructed images using the peak signal-to-noise ratio (PSNR) in RGB colorspace. 
The rate is measured by bits per pixel (bpp) for evaluation and bits per dimension (bpd) for training.
The distortion metrics $d$ and $d_1$ in Equation~\ref{eq:loss-msvq} are the mean squared error (MSE).
We also use BD-rate/BD-PSNR~\cite{BD-metric} to calculate the average rate-distortion performance gain. 

\textbf{Implementation details.}
Our image codec consists of 16 quantization layers, where $(L_1, L_2, L_3) = (6, 6, 4)$.
The detailed configurations are described in the supplementary.
We train six models with $\lambda \in \{64, 128, 256, 512, 1024, 2048\}$ for different RD points. $\beta$ is set equivalent to $\lambda$. 
We initialize the models using Algorithm~\ref{alg:MSVQinit} with 0.6M iterations and then optimize the full models with 1.7M iterations.
We use the Adam optimizer~\cite{Adam} with a batch size of 8. 
The initial learning rate is set as $10^{-4}$, which is reduced to $10^{-5}$ for the final 0.2M iterations.

\subsection{Performance}
\textbf{RD curves.}
 We compare our method with two traditional coding methods: BPG and VVC~\cite{VVC} (VTM-18.2), and six learning-based methods including Zou-CVPR22~\cite{Zou-CVPR2022}, Zhu-ICLR22~\cite{Zhu-ICLR2022}, Zhu-CVPR22~\cite{Zhu-CVPR2022}, Cheng-CVPR20~\cite{Cheng-CVPR2020}, Balle-ICLR18~\cite{Balle-ICLR2018} and Agustsson-NIPS17~\cite{Agustsson-NIPS2017}.
Among them, Zhu-CVPR22~\cite{Zhu-CVPR2022} and Agustsson-NIPS17~\cite{Agustsson-NIPS2017} are two latent-space VQ methods. 
As shown in Figure~\ref{fig:rd_kodak_clic_psnr}, our method is on par with previous methods on Kodak but surpasses them on CLIC 2021 validset. 
Even if our method is trained on the COCO dataset with a resolution of about 600 $\times$ 400, the generalization performance on the high-resolution CLIC dataset is superior to the previous latent-space VQ method~\cite{Zhu-CVPR2022}.

\begin{table}[th]
\begin{center}
\begin{tabular}{crrrrrr}
\hline
\multirow{2}{*}{}
& \multicolumn{5}{c}{Time (ms)} \\ \cline{2-7}
&  &  NVT & CEM & VQ & EC & Total \\ \hline
\multirow{2}{*}{Kodak} & Encode & 48 & 30 & 12 & 24 & 114 \\ 
& Decode & 23 & 13 & 4 & 28 & 68 \\ \cline{1-7}

\multirow{2}{*}{CLIC} & Encode & 270 & 121 & 63 & 89 & 543 \\ 
& Decode & 154 & 38 & 14 & 101 & 307 \\ \cline{1-7}
\end{tabular}
\end{center}
\vspace{-4mm}
\caption{Encoding \& decoding time. 
NVT refers to the nonlinear vector transform including VT Units and Downscale/Upscale layers. 
CEM is the conditional entropy model. 
EC refers to entropy encoding/decoding with one CPU core. GPU is GTX 1080Ti.}
\label{tab:complexity}
\vspace{-2mm}
\end{table}

\textbf{BD-rate vs. model complexity.}
In Figure~\ref{fig:rdc_clic_psnr}, we compare the BD-rate, decoding time, and model parameters of different methods on CLIC 2021 validset.
The BD-rate is computed over VTM 18.2, and the reported time is the average decoding time per image across the entire dataset. 
For a fair comparison, we evaluate the model complexity at high rate points (larger than 0.5 bpp on CLIC) except for Zhu-CVPR22\footnote{The authors only release one low-rate model. Therefore we report the estimated results based on the  description in their paper.}~\cite{Zhu-CVPR2022} and Guo-TCSVT21\footnote{The results are provided by the authors.}~\cite{Guo-TCSVT2021}.
All methods run with one GTX 1080Ti GPU and one CPU core, based on their official code or public reimplementations~\cite{CompressAI}. 
It can be observed that our method has the best rate-distortion-complexity performance. 
\textbf{More importantly}, compared to the recent VQ-based approach Zhu-CVPR2022~\cite{Zhu-CVPR2022} that relies on computationally expensive scalar transform and unconstrained VQ, our method has a significantly reduced model size (about $5\times$ smaller), faster decoding speed and better RD performance. 
The BD-rate of our method against Zhu-CVPR2022 is -10.9\% on CLIC.

In Table~\ref{tab:complexity}, we provide detailed results regarding the encoding/decoding time of our method. 
 It is worth noting that the encoding time is approximately twice the decoding time, which mainly stems from two factors: 
1) the encoder needs to run with most of the VT units, where the decoder only has to execute the decoder VT units; 
2) The computational gap between ECVQ encoding and decoding. 
We provide a more detailed explanation in the supplementary.

\subsection{Ablation Study and Analysis}

\begin{table}[h]
\begin{center}
\begin{tabular}{lccc}
\hline
\multirow{2}{*}{} & \multirow{2}{*}{BD-rate} &
Decoding & Model \\ 
& & time (ms) & parameters (M) \\ \hline
Full model        & 0.0\% & 307 & 12.8 \\
A1: w/o VT   & 7.3\% & 405 & 49.6 \\ 
A2: w/o ECVQ & 23.9\% & 268 & 11.5 \\
A3: w/o init & 17.1\% & 307 & 12.8 \\

\hline
\end{tabular}
\end{center}

\vspace{-4mm}
\caption{Ablation study on CLIC 2021 validset. BD-rate is computed over the full models (smaller is better). }
\label{tab:ablation}
\vspace{-2mm}
\end{table}

\textbf{Vector transform.}
To investigate the impact of the proposed VT, we replace the VT unit with the convolutional resblock used in previous image coding methods~\cite{Cheng-CVPR2020, Zhu-CVPR2022}. 
The result in Table~\ref{tab:ablation} (``A1: w/o VT'') shows that VT provides a better decorrelation ability with less complexity.

\textbf{Latent-space ECVQ.}
As mentioned before, compared to ECVQ, the unconstrained VQ is theoretically suboptimal in source space. 
To verify the effectiveness of ECVQ in latent space, we replace ECVQ with the unconstrained VQ in previous works~\cite{Zhu-CVPR2022, Agustsson-NIPS2017}, and retrain our model.
We manually adjust the codebook size to control the rate-distortion trade-off.
As shown in Table~\ref{tab:ablation} (``A2: w/o ECVQ''), unconstrained VQ  brings a significant performance drop.

\textbf{Progressive initialization.} 
To evaluate the impact of the progressive initialization in Algorithm~\ref{alg:MSVQinit}, we optimize the full model with randomly initialized parameters. 
As shown in Table~\ref{tab:ablation} (``A3: w/o init''), random initialization causes a large performance drop. 
We find that only a part of codewords is activated for quantization, and the other codewords are ignored by ECVQ due to their extremely low probability. ECVQ with random initialization is more likely to make codewords ``dead'', restricting its performance at high rate.

\begin{figure}[t]
 \centering
 \begin{subfigure}{\linewidth}
\includegraphics[width=\linewidth, clip]{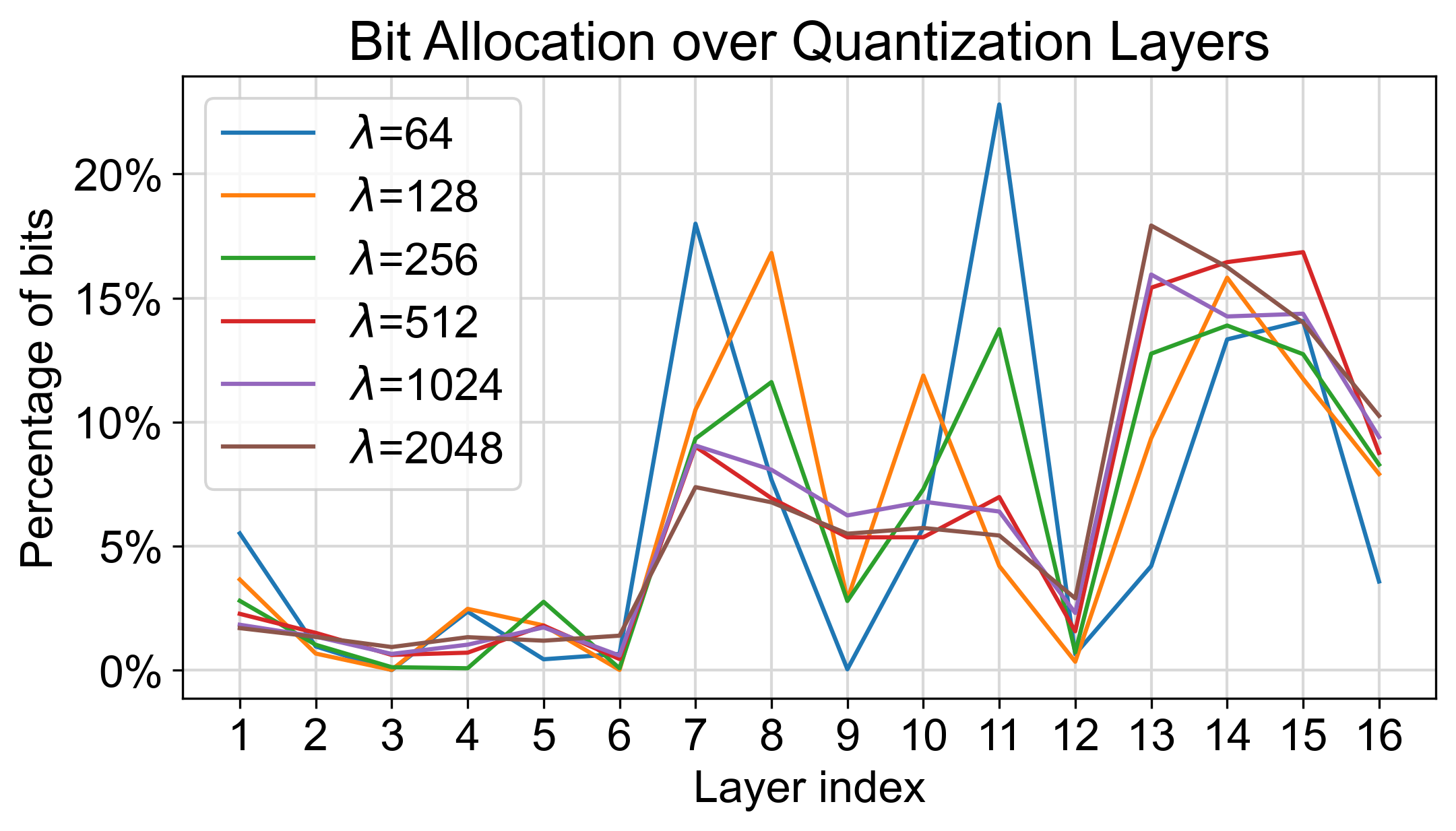}
 \end{subfigure}

\vspace{-4mm}
 \caption{Layer-adaptive bit allocation results. We report the average percentage of bits of different quantization layers on Kodak. }
\label{fig:ba_kodak_layerwise}
\vspace{-4mm}
\end{figure}

\textbf{Layer-adaptive bit allocation.} 
Note that our model consists of 16 quantization layers, which are joint optimized for the target RD trade-off. 
To investigate the bit allocation of our model, we show the percentage of bits of each layer in Figure~\ref{fig:ba_kodak_layerwise}.  
It can be observed that the model allocates more bits to high-resolution layers as $\lambda$ increases.

For more results, analysis and visualization, please refer to the supplementary.

\section{Conclusion}
In this paper, we demonstrate the superiority of VQ over SQ with nonlinear transform based on empirical results on some toy sources.
We propose a novel VQ-based coding scheme named nonlinear vector transform coding to make better use of VQ while keeping complexity low.
Specifically, three technical points are proposed, including 1) a multi-stage quantization strategy, 2) nonlinear vector transforms and 3) RD optimization with entropy-constrained VQ in latent space.
Our method and these analyses shed light in utilizing VQ for better neural network-based compression.

\section*{Acknowledgments}
This work was supported in part by NSFC under Grant U1908209, 62021001.

{\small
\bibliographystyle{ieee_fullname}
\bibliography{egbib}
}

\clearpage
\appendix
\section*{A. On Toy Sources }
\begin{figure*}[th]
\centering
\begin{subfigure}{0.24\linewidth}
\includegraphics[width=\linewidth, trim=0cm 0cm 0cm 0cm]{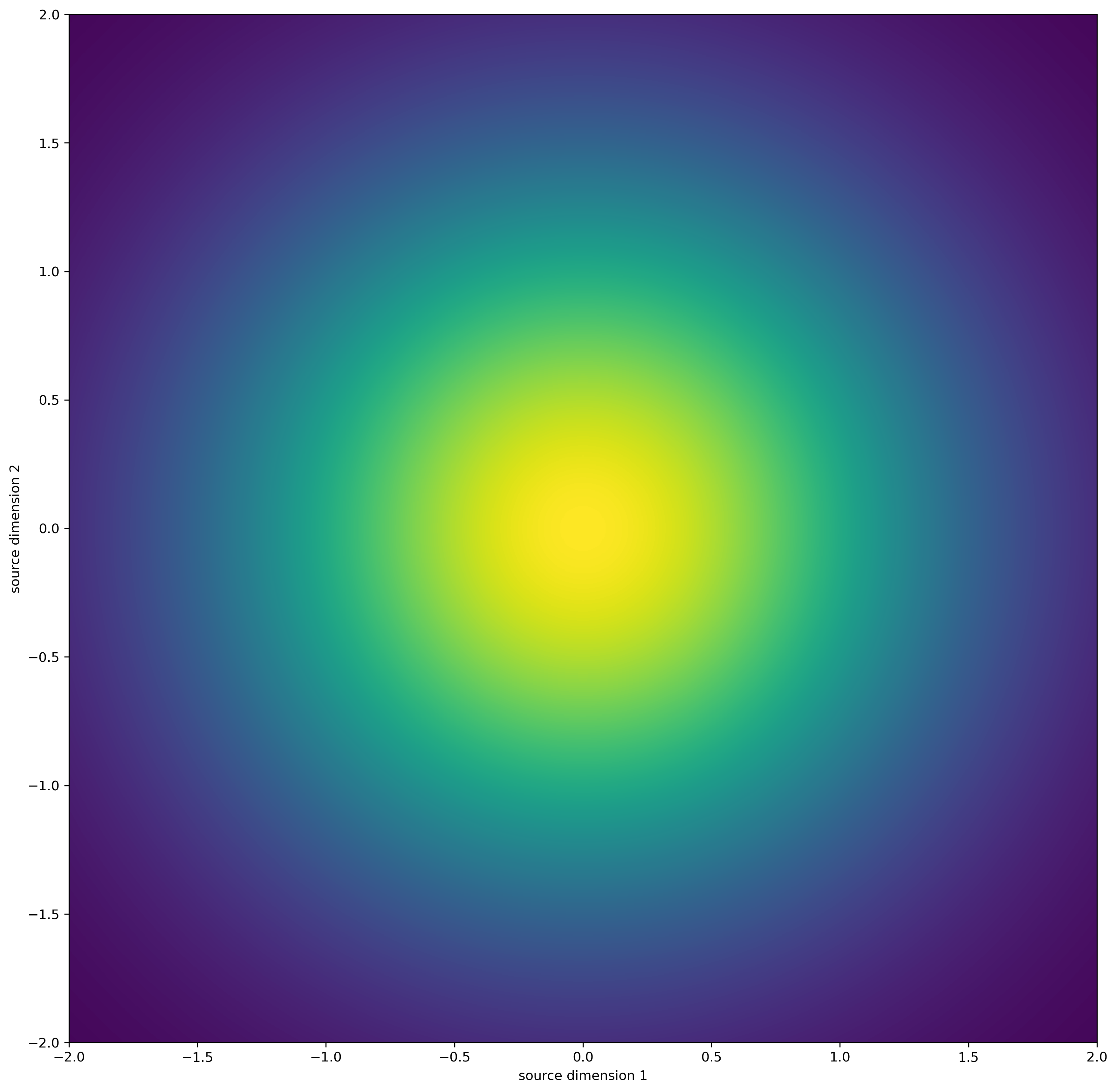}
\caption{Isotropic Gaussian}
\end{subfigure}
\begin{subfigure}{0.24\linewidth}
\includegraphics[width=\linewidth, trim=0cm 0cm 0cm 0cm]{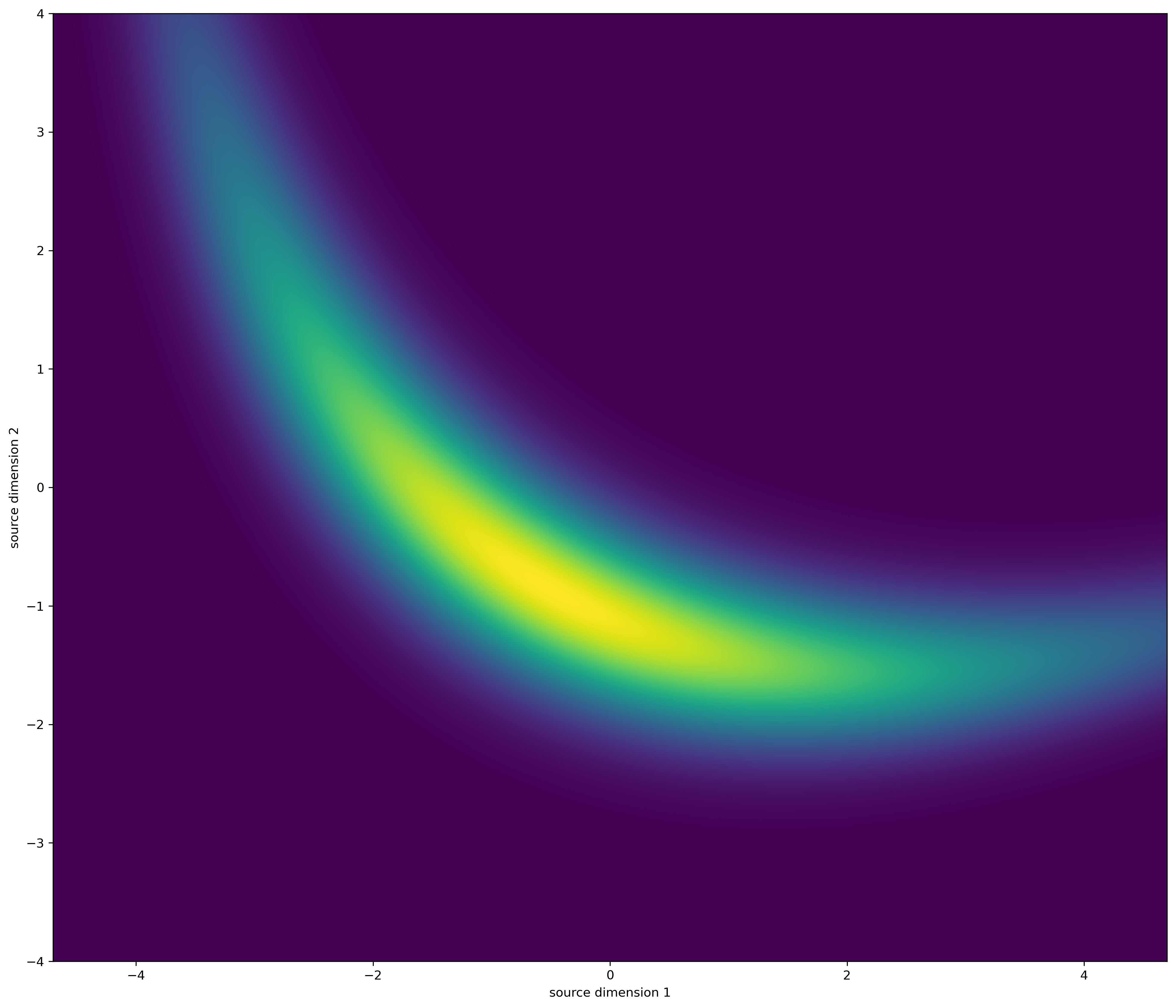}
\caption{Banana}
\end{subfigure}
\begin{subfigure}{0.24\linewidth}
\includegraphics[width=\linewidth, trim=0cm 0cm 0cm 0cm]{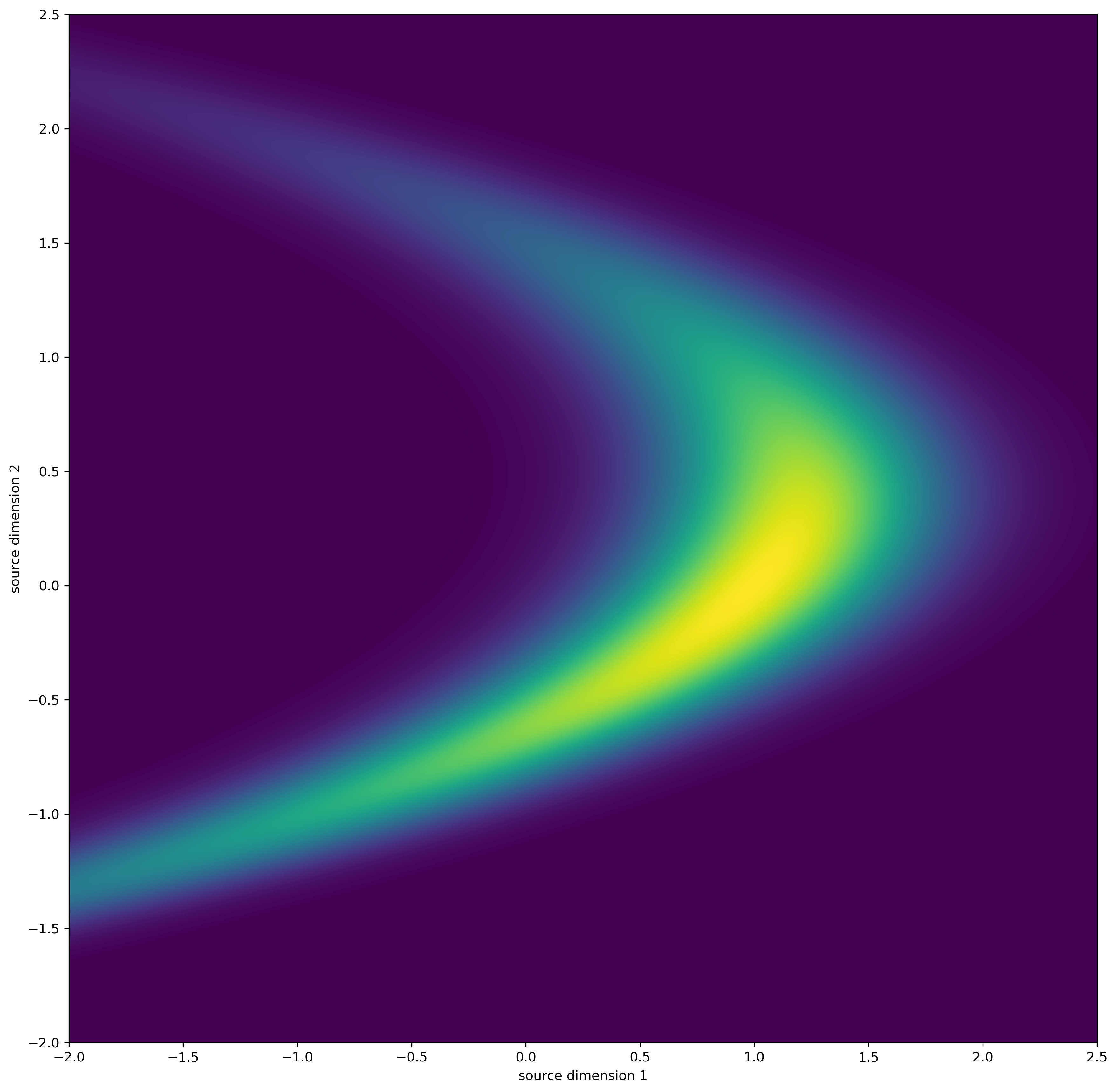}
\caption{Boomerang}
\end{subfigure}
\begin{subfigure}{0.24\linewidth}
\includegraphics[width=\linewidth, trim=0cm 0cm 0cm 0cm]{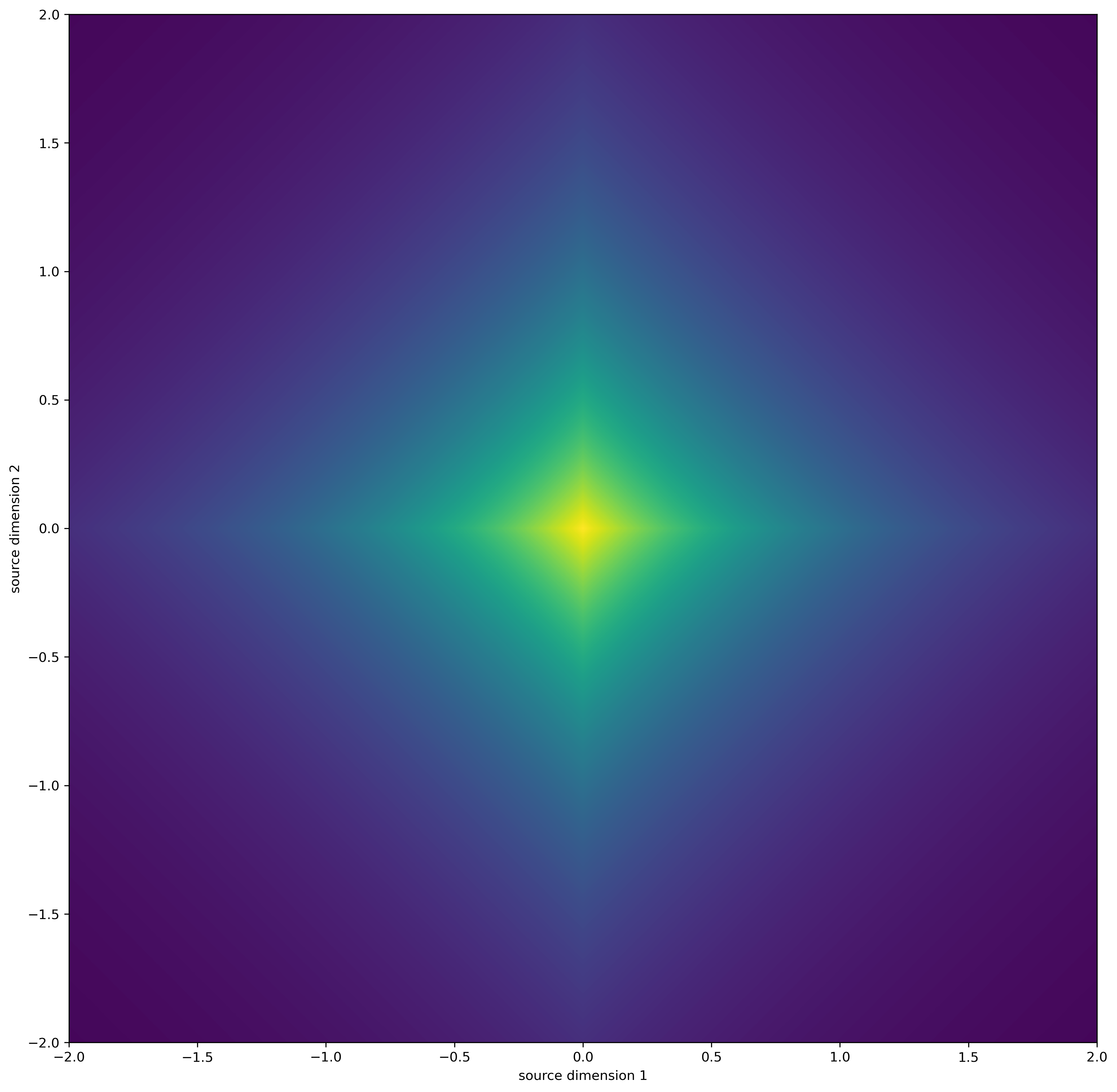}
\caption{Laplace}
\end{subfigure}

\begin{subfigure}{0.24\linewidth}
\includegraphics[width=\linewidth, trim=0cm 0cm 0cm 0cm]{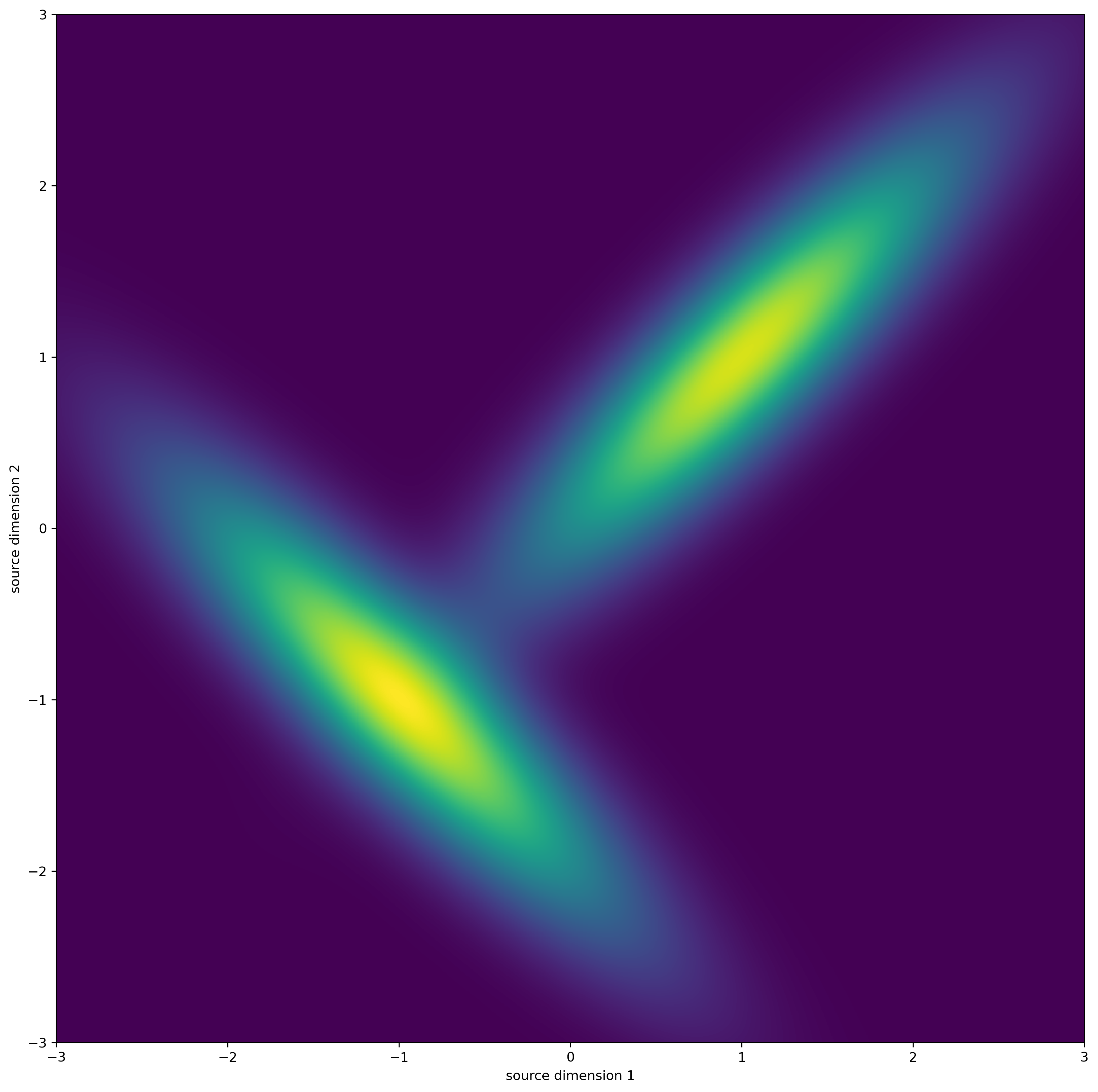}
\caption{Gaussian Mixture}
\end{subfigure}
\begin{subfigure}{0.24\linewidth}
\includegraphics[width=\linewidth, trim=0cm 0cm 0cm 0cm]{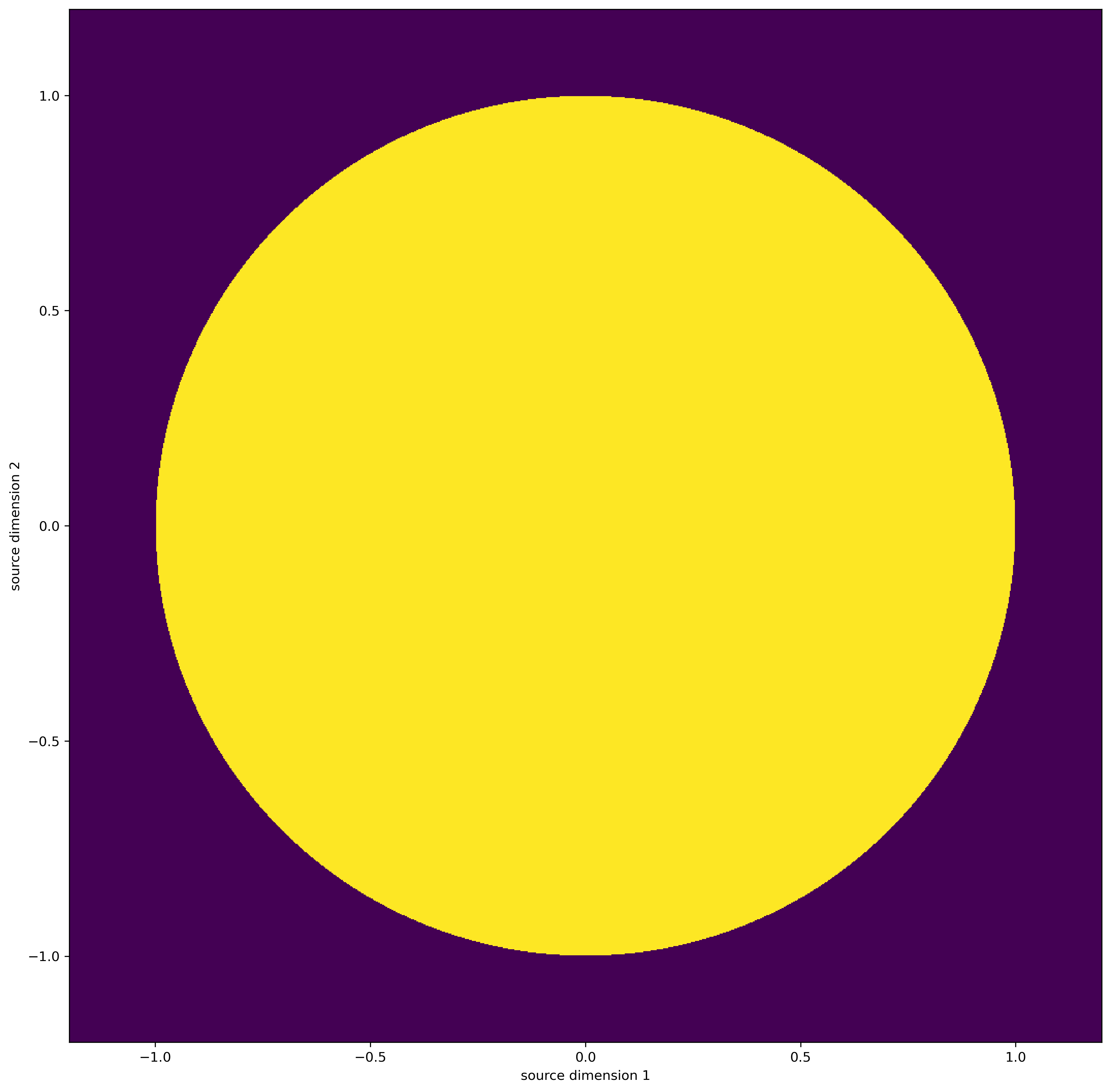}
\caption{Shpere0}
\end{subfigure}
\begin{subfigure}{0.24\linewidth}
\includegraphics[width=\linewidth, trim=0cm 0cm 0cm 0cm]{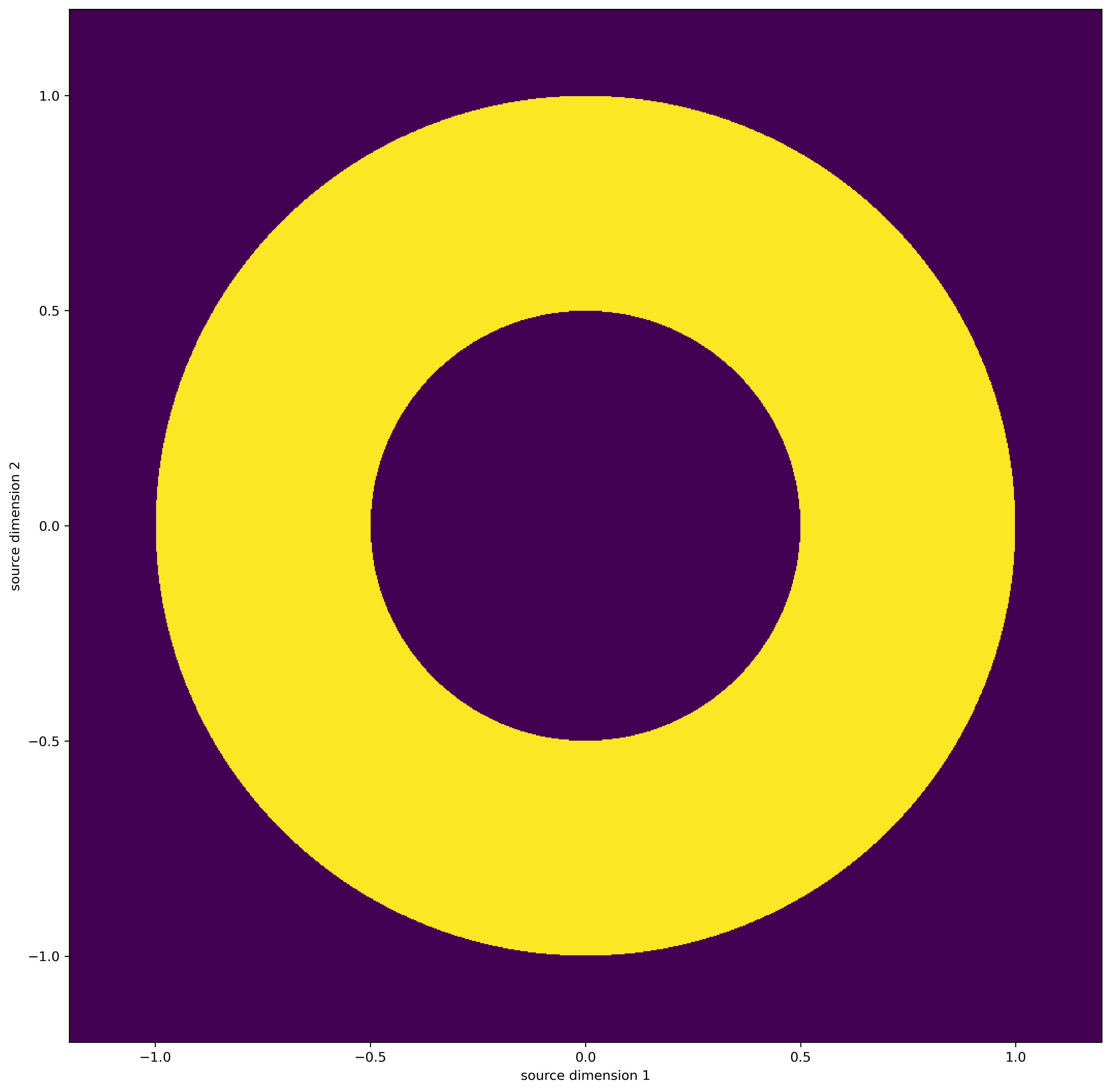}
\caption{Shpere50}
\end{subfigure}
\begin{subfigure}{0.24\linewidth}
\includegraphics[width=\linewidth, trim=0cm 0cm 0cm 0cm]{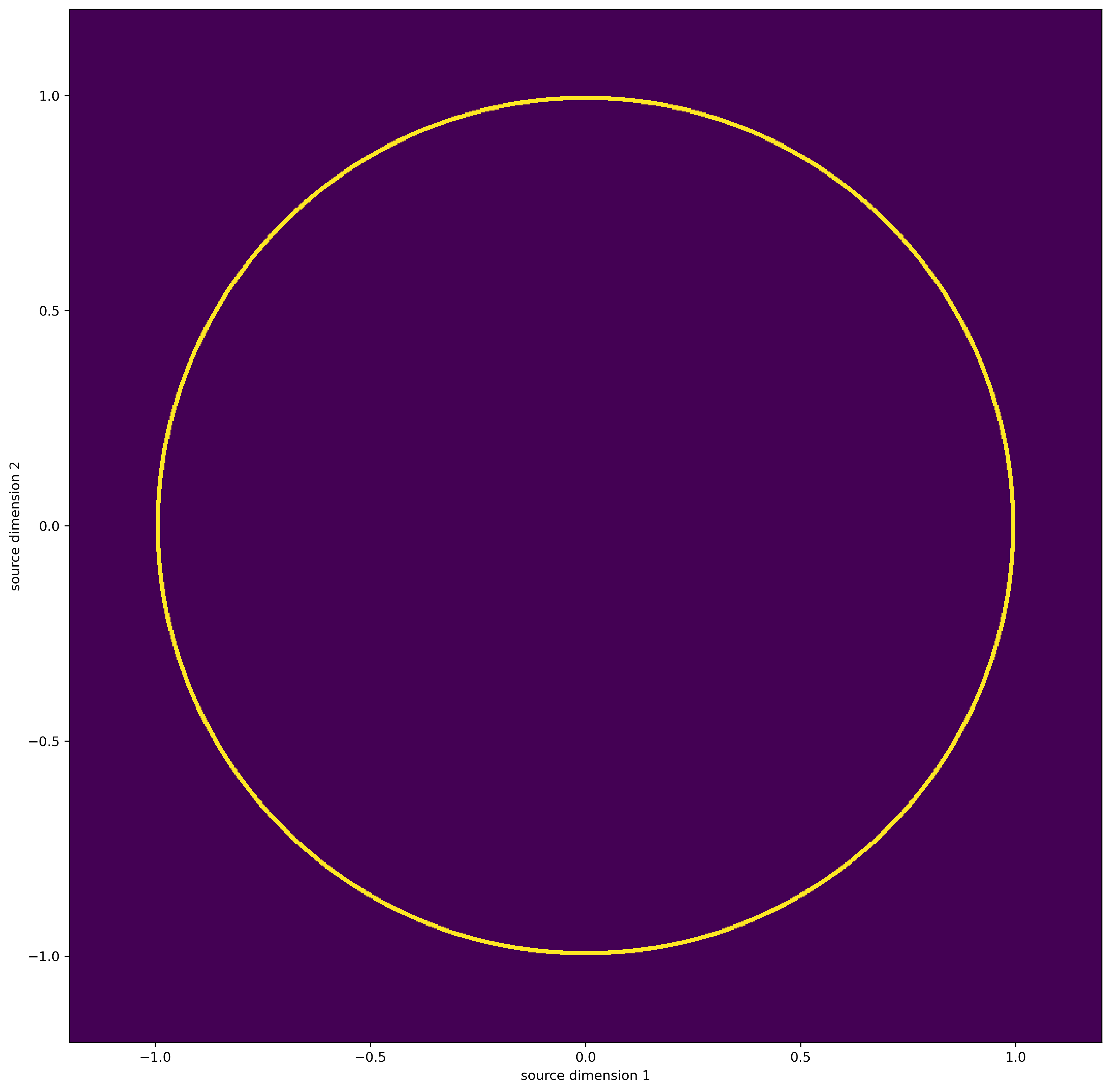}
\caption{Shpere99}
\end{subfigure}
\caption{Visualization of 2-d sources. Yellow means high probability density and purple means low probability density.}
\label{fig:vis_toy_sources}
\end{figure*}

We investigate the advantages of ECVQ on 11 different source distributions with varied dimensions. 
Among them, we show the 8 two-dimentional distributions in Figure~\ref{fig:vis_toy_sources}, including Isotropic Gaussian, Banana, Boomerang, Laplace, Gaussian Mixture, Sphere0, Shpere50 and Shpere99.

The details about the architecture of codecs are described in Section \hyperref[sec:a.1]{A.1}.
The experimental results and analysis are provided in Section \hyperref[sec:a.2]{A.2}.

\subsection*{A.1. Architecture}\label{sec:a.1}
We implement three different codecs for comparison: VQ/ECVQ, SQ with nonlinear transform (NTC), and VQ/ECVQ with nonlinear transform (NT-VQ/NT-ECVQ). 
\paragraph{VQ/ECVQ.}
As mentioned in the main paper, the encoder function of \textbf{VQ} with dimension $k$ and size $N$ is as follows:

\begin{equation}
\begin{aligned}
i = \mathop{\arg\min}_i d(\boldsymbol{x}, \boldsymbol{c}_i),
\end{aligned}
\label{eq:}
\end{equation}
where $d$ is a distortion metric and $\boldsymbol{c}_i$ is a codeword of the codebook $\boldsymbol{c}=\{\boldsymbol{c}_i \in \mathbb{R}^k| i=0, 1, ..., N-1\}$.
The decoder function is a lookup operation: $\boldsymbol{\hat{x}}=\boldsymbol{c}_i$.
The codeword probability $P(i)=p_i$ is parameterized with the Softmax function and unnormalized logits $\boldsymbol{w}=(w_1, w_2, ..., w_N)$:
\begin{equation}
\begin{aligned}
p_i = \frac{e^{-w_i}}{\sum_{j=1}^N e^{-w_j}}.
\end{aligned}
\label{eq:UEM}
\end{equation}
The rate loss is $\mathbb{E}_{\boldsymbol{x}}{\left[-\log{p_i}\right]}$ and the distortion loss is $\mathbb{E}_{\boldsymbol{x}} {\left[ d(\boldsymbol{x}, \boldsymbol{\hat{x}})\right]}$. 
The encoder function of \textbf{ECVQ} is improved as follows:
\begin{equation}
\begin{aligned}
i = \mathop{ \arg\min}_i \left[ -\log{p_i} + \lambda d(\boldsymbol{x}, \boldsymbol{c}_i)\right],
\end{aligned}
\label{eq:}
\end{equation}
where $\lambda$ controls the RD trade-off.

\begin{figure*}[th]
 \centering
  \begin{subfigure}{\linewidth}
 \includegraphics[width=\linewidth, trim=0cm 0cm 0cm 0cm]{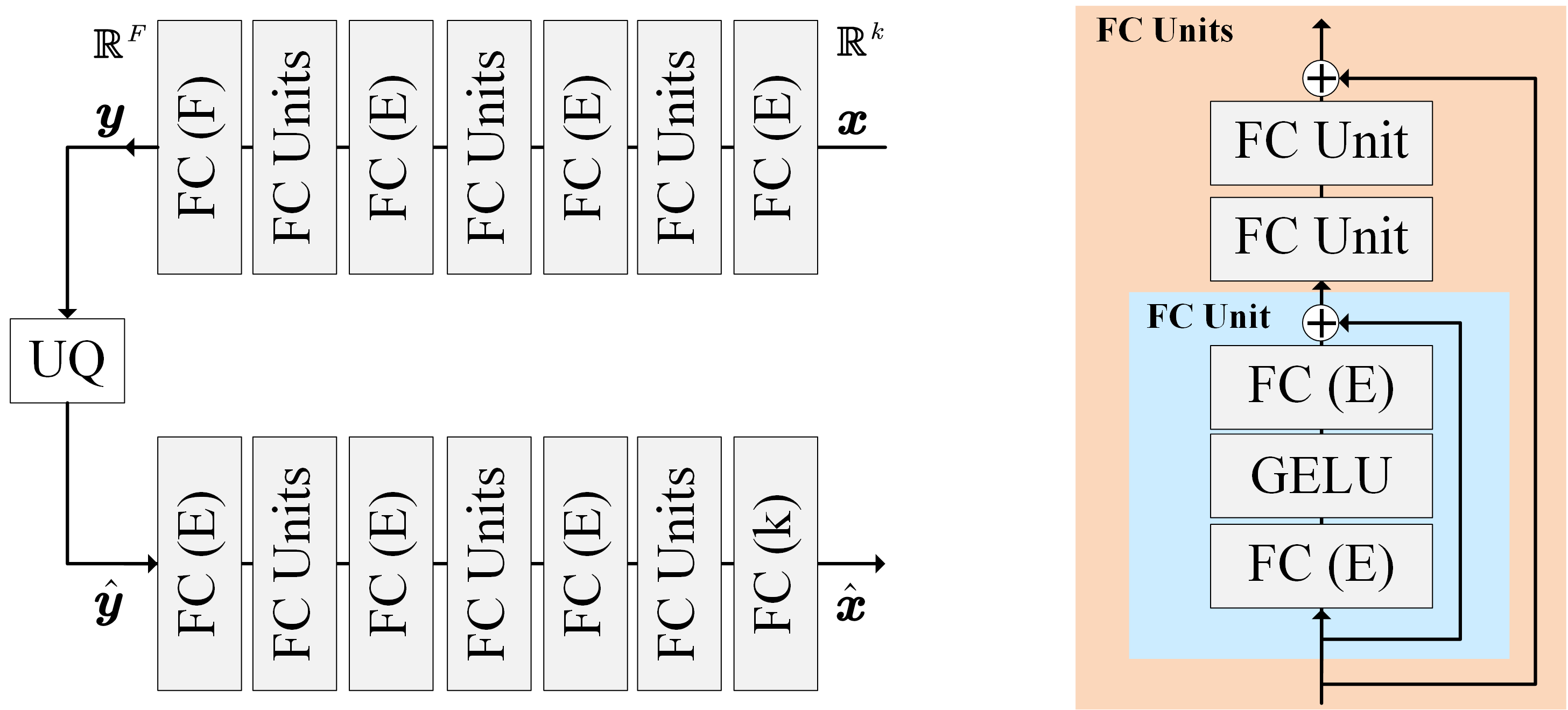}
 \end{subfigure}
\caption{Architecture of NTC on toy sources. ``FC (E)'' refers to a fully-connected layer with $E$ output channels. $F$ is equal to the data dimension $k$. $E=128$ for 2-d distributions, and $E=384$ for 4-d, 8-d and 16-d distributions. ``UQ'' is the uniform scalar quantization. ``GELU'' is the Gaussian Error Linear Units~\cite{GELU}.}
\label{fig:arch_ntc}
\end{figure*}

\paragraph{NTC.}
The architecture of NTC for toy sources is shown in Figure~\ref{fig:arch_ntc}. Input $\boldsymbol{x}$ is transformed into latents $\boldsymbol{y}=g_a(\boldsymbol{x})$ by a analysis transform $g_a$, which is then quantized by a uniform scalar quantizer (rounding to integers). 
The quantized latent $\boldsymbol{\hat{y}}=\lfloor \boldsymbol{y} \rceil$ is then mapped back to the reconstruction $\boldsymbol{\hat{x}}=g_s(\boldsymbol{\hat{y}})$ by a synthesis transform $g_s$. In training, the rounding operation approximated by adding uniform noise~\cite{Balle-ICLR2018, NTC}. We use the factorized prior~\cite{Balle-ICLR2018} for entropy modeling.
The overall loss is as follows:

\begin{equation}
\begin{aligned}
\mathcal{L} =  \mathbb{E}_{\boldsymbol{x}}{\left[-\log{P(\boldsymbol{\hat{y}})} \right]} + \lambda \mathbb{E}_{\boldsymbol{x}}{d(\boldsymbol{x}, \boldsymbol{\hat{x}})}.
\end{aligned}
\label{eq:}
\end{equation}

\paragraph{NT-VQ/NT-ECVQ.}
To verify the effectiveness of ECVQ in latent space, we replace the uniform scalar quantization of NTC with VQ/ECVQ, named NT-VQ/NT-ECVQ.
For \textbf{NT-VQ}, the encoding function is as follows:
\begin{equation}
\begin{aligned}
i = \mathop{\arg\min}_i d_1(g_a(\boldsymbol{x}), \boldsymbol{c}_i),
\end{aligned}
\label{eq:}
\end{equation}
and the encoding function for \textbf{NT-ECVQ} is:
\begin{equation}
\begin{aligned}
i = \mathop{ \arg\min}_i \left[ -\log{p_i} + \lambda d_1(g_a(\boldsymbol{x}), \boldsymbol{c}_i)\right].
\end{aligned}
\label{eq:}
\end{equation}
We have $\boldsymbol{y} = g_a(\boldsymbol{x})$, $\boldsymbol{\hat{y}} = \boldsymbol{c}_i$ and $\boldsymbol{\hat{x}} = g_s(\boldsymbol{\hat{y}})$.
To jointly optimize the transform and VQ codebook, we employ the straight-through estimator (STE) used in previous works~\cite{VQVAE, Agustsson-NIPS2017}:
\begin{equation}
\begin{aligned}
\frac{d\boldsymbol{\hat{y}}}{d\boldsymbol{y}} = 1,
\end{aligned}
\label{eq:}
\end{equation}
and propose a new rate-distortion loss for latent space VQ:
\begin{equation}
\begin{aligned}
\mathcal{L} =  \mathbb{E}_{\boldsymbol{x}}{\left[-\log{p_i} \right]} + \lambda \mathbb{E}_{\boldsymbol{x}}{d(\boldsymbol{x}, \boldsymbol{\hat{x}})} + \beta \mathbb{E}_{\boldsymbol{x}}{d_1(\boldsymbol{y}, \boldsymbol{c}_i)},
\end{aligned}
\label{eq:loss-ntecvq}
\end{equation}
where $d$ is the source-space distortion metric, and $d_1$ is the latent-space VQ distortion metric. 
$\lambda$ controls the rate-distortion trade-off. 
$\beta$ controls the trade-off between $d$ and $d_1$.
In practice, both $d$ and $d_1$ are measured by MSE, and $\beta$ is equal to $\lambda$.

\subsection*{A.2. Experiments}\label{sec:a.2}
\begin{figure*}[th]
\centering
\begin{subfigure}{\linewidth}
\includegraphics[width=0.33\linewidth, trim=0cm 0cm 0cm 0cm]{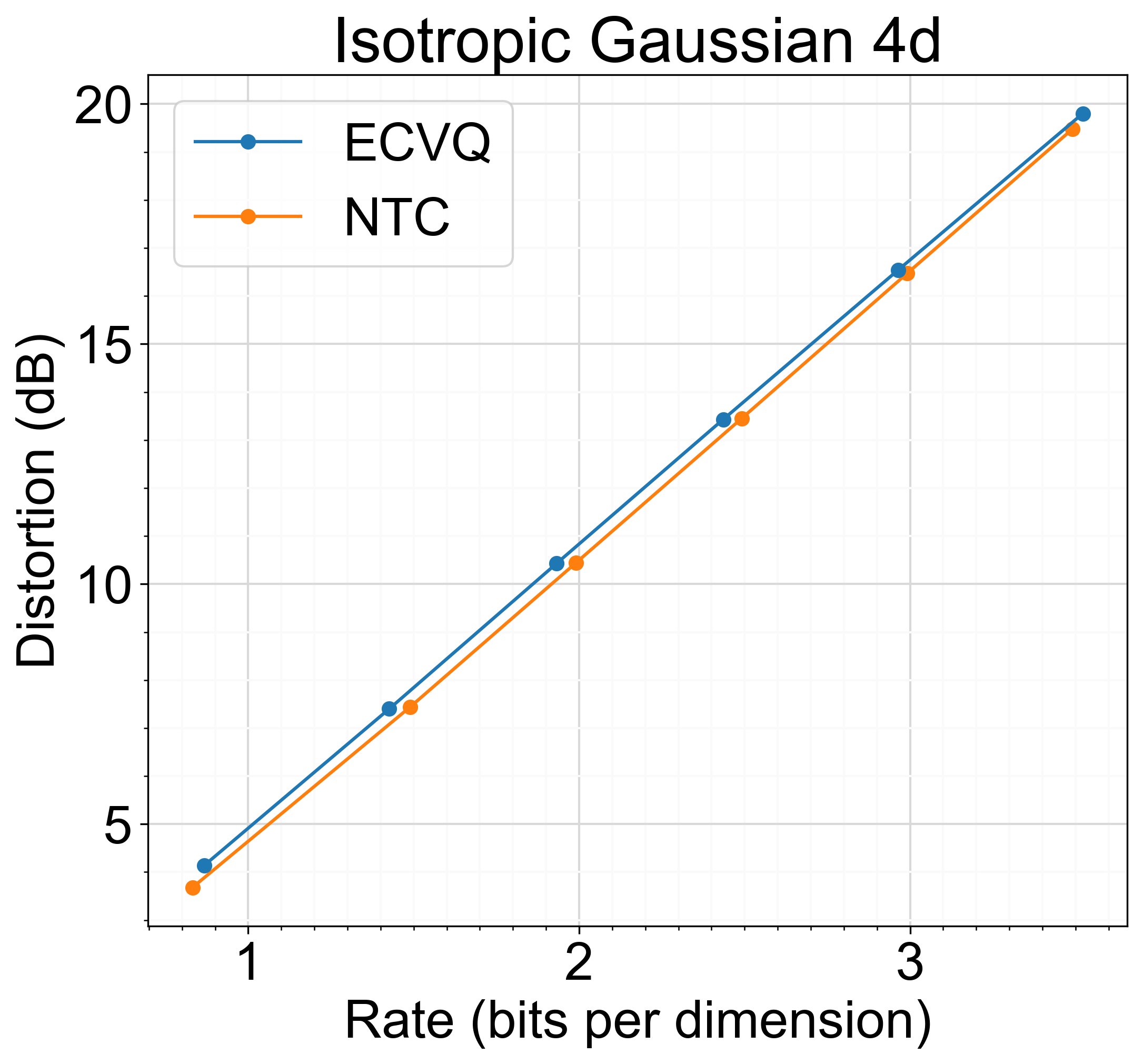}
\includegraphics[width=0.33\linewidth, trim=0cm 0cm 0cm 0cm]{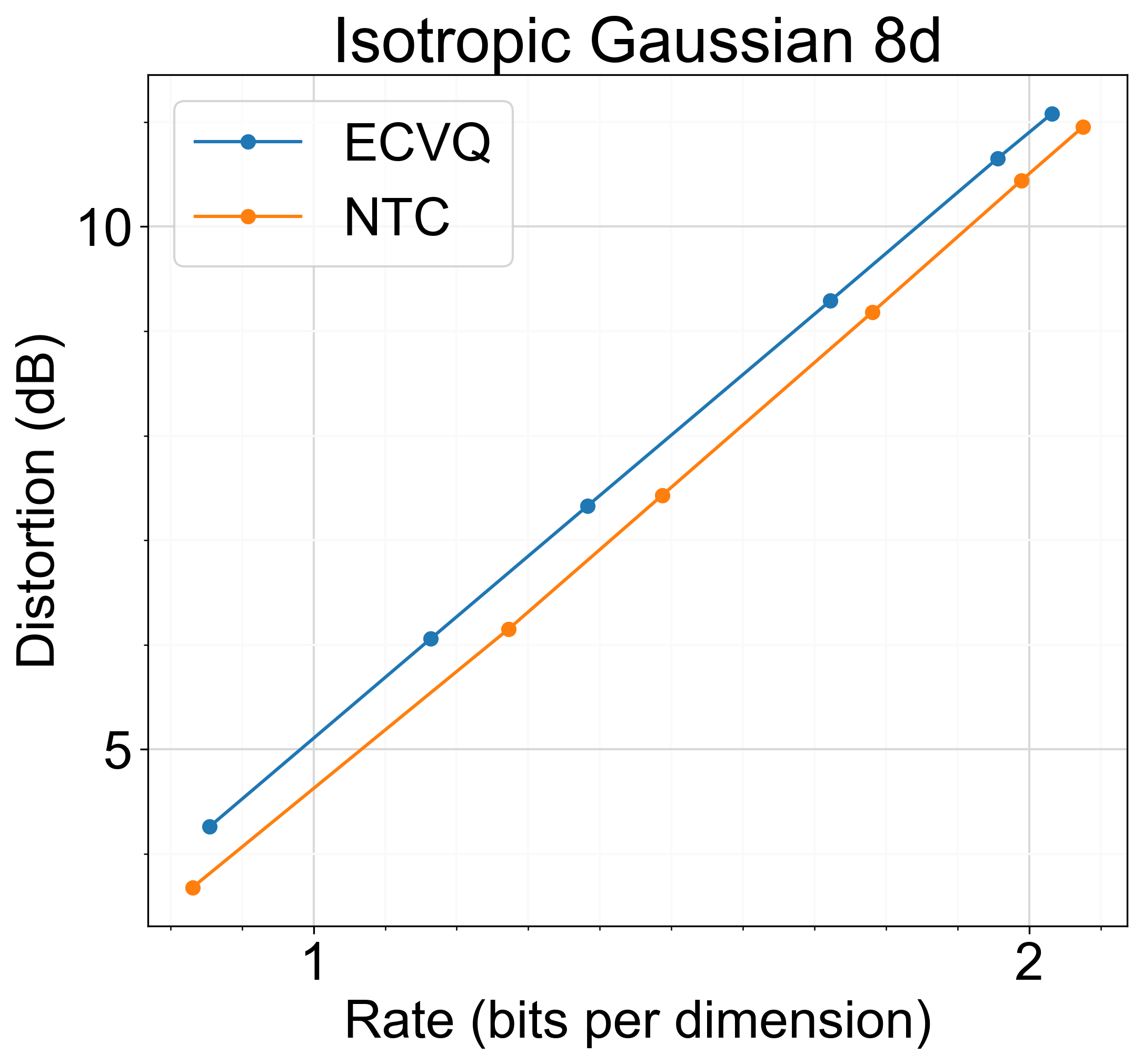}
\includegraphics[width=0.33\linewidth, trim=0cm 0cm 0cm 0cm]{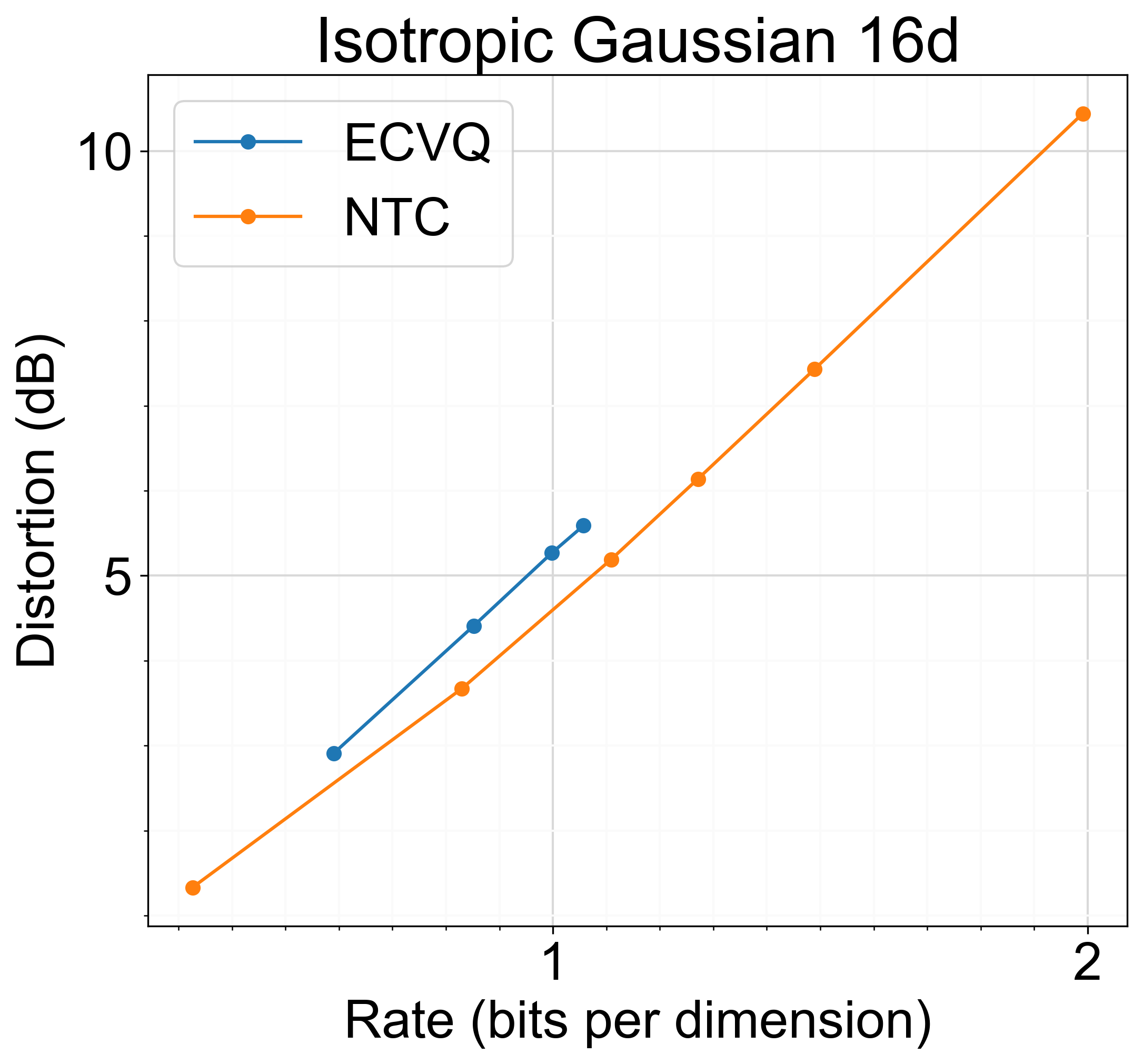}
\end{subfigure}
\vspace{2mm}
\begin{subfigure}{\linewidth}
\includegraphics[width=0.33\linewidth, trim=0cm 0cm 0cm 0cm]{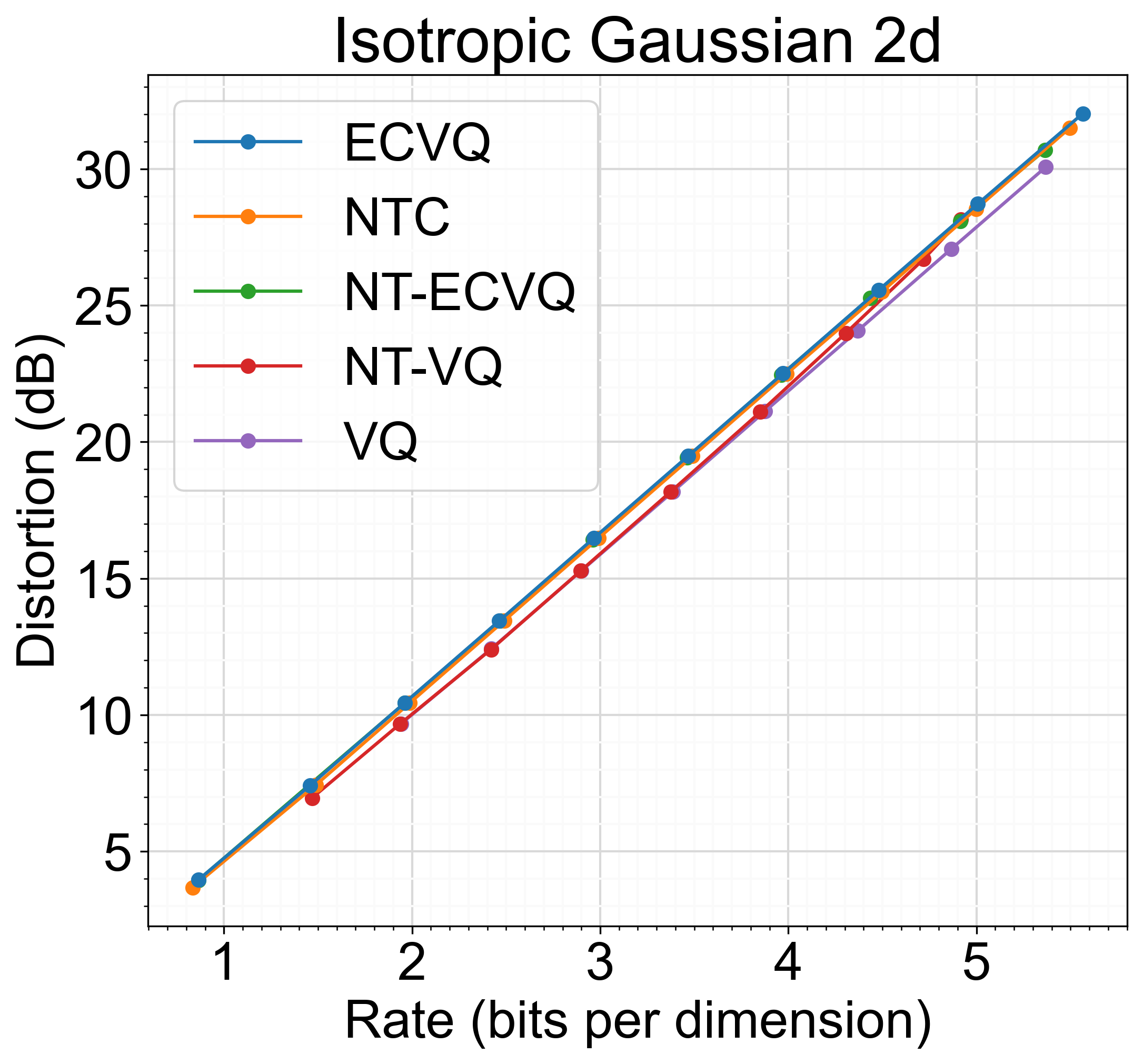}
\includegraphics[width=0.33\linewidth, trim=0cm 0cm 0cm 0cm]{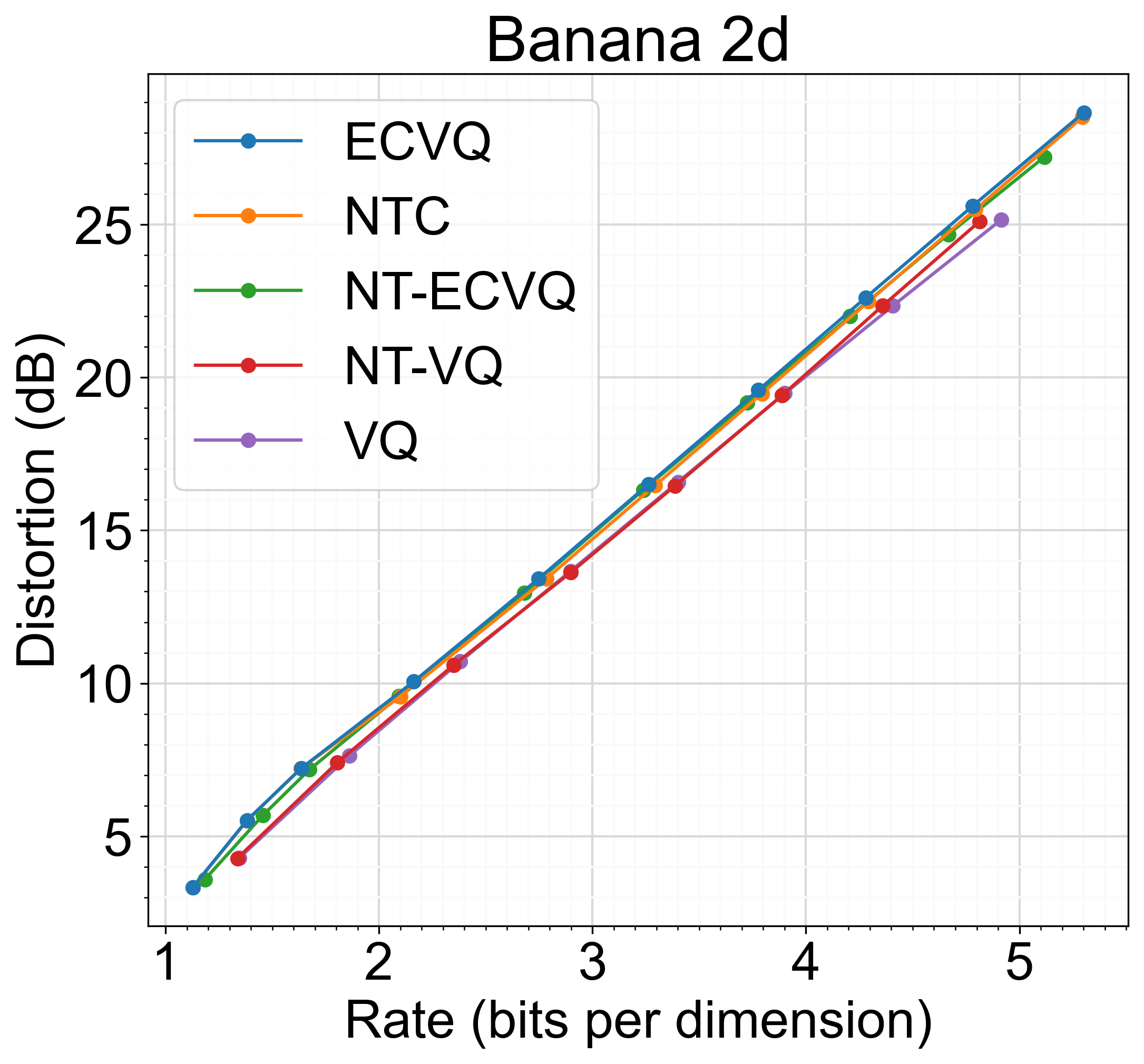}
\includegraphics[width=0.33\linewidth, trim=0cm 0cm 0cm 0cm]{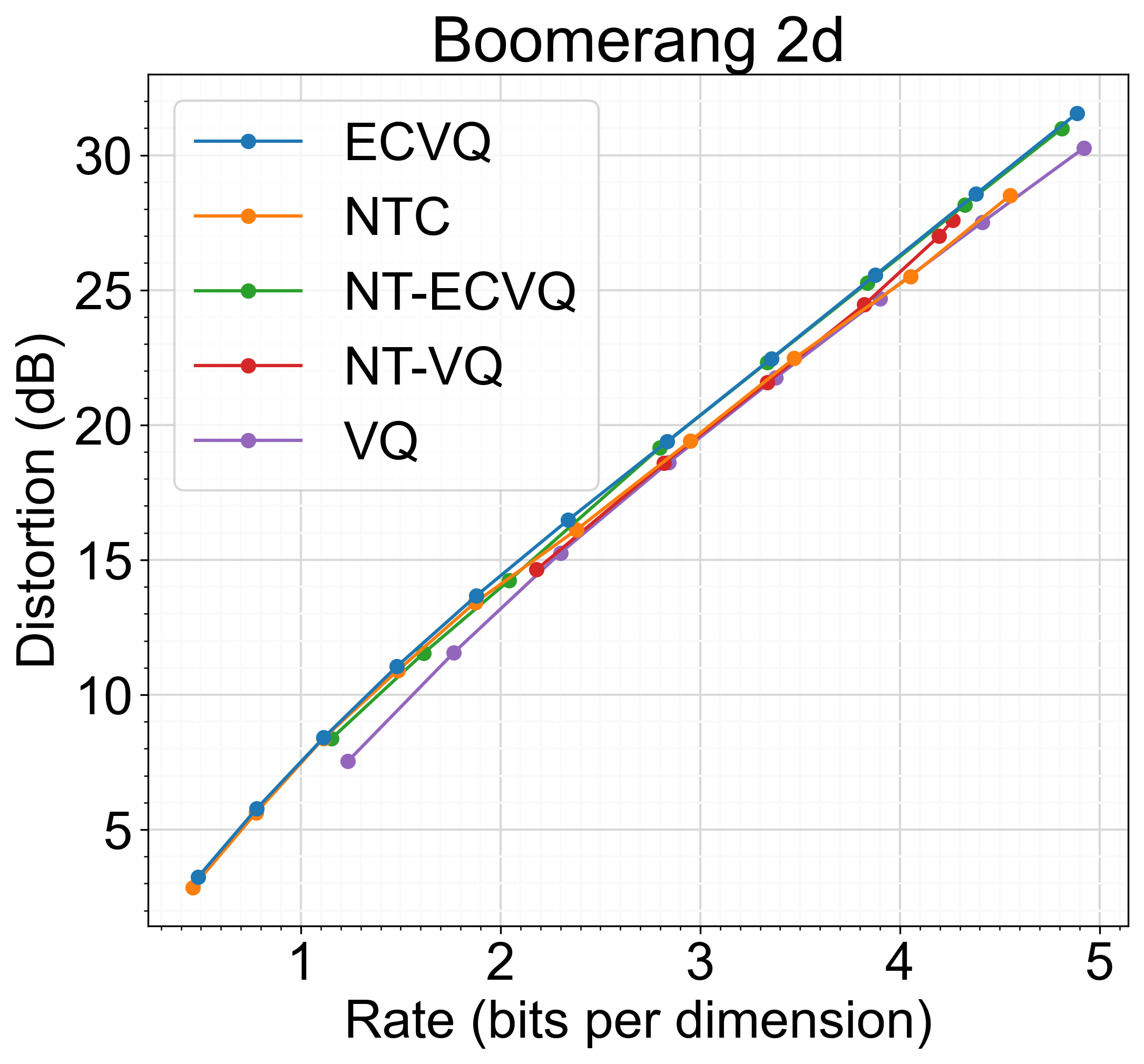}
\end{subfigure}
 \vspace{2mm}
\begin{subfigure}{\linewidth}
\includegraphics[width=0.33\linewidth, trim=0cm 0cm 0cm 0cm]{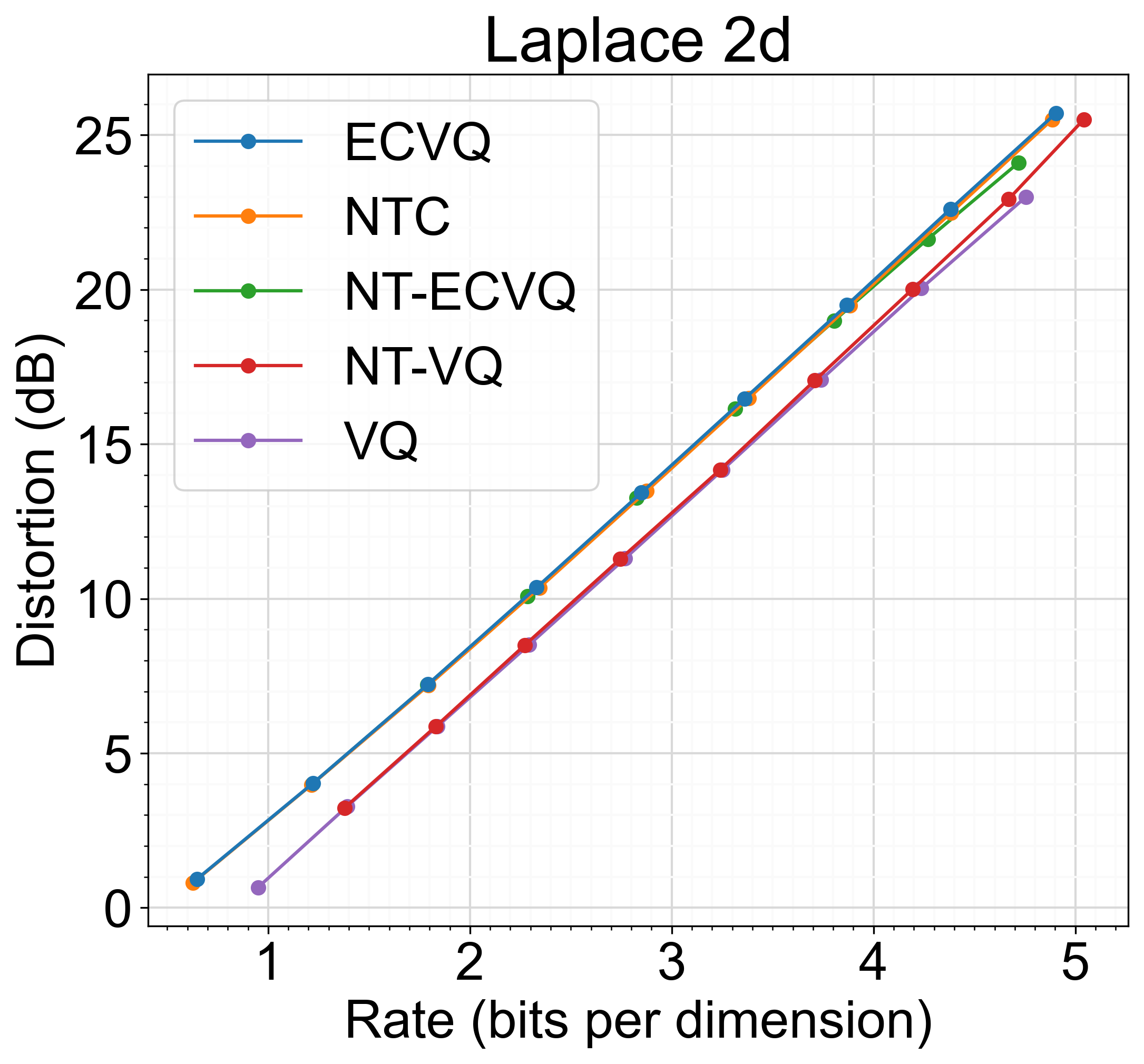}
\includegraphics[width=0.33\linewidth, trim=0cm 0cm 0cm 0cm]{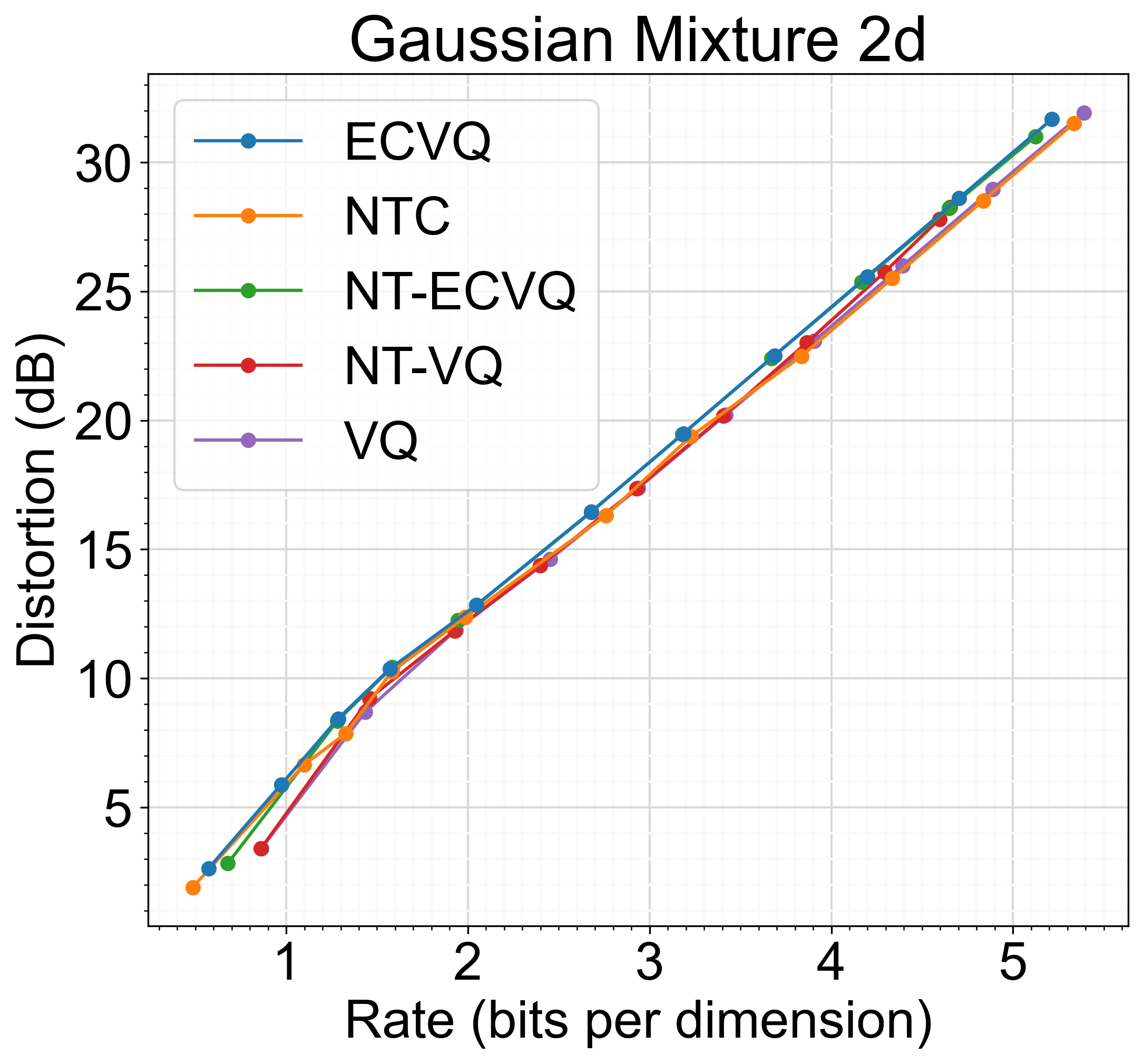}
\end{subfigure}
 \vspace{2mm}
\begin{subfigure}{\linewidth}
\includegraphics[width=0.33\linewidth, trim=0cm 0cm 0cm 0cm]{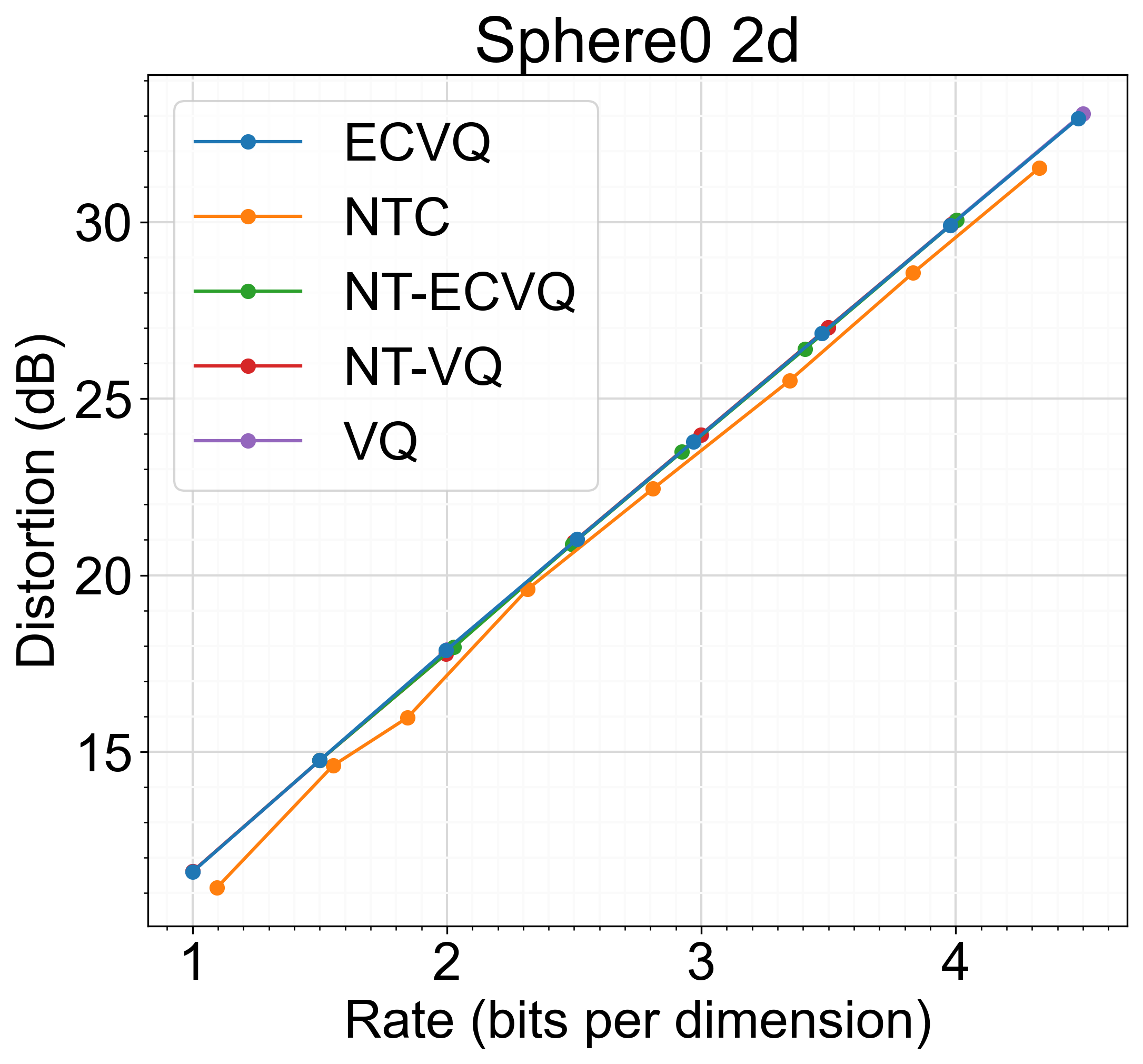}
\includegraphics[width=0.33\linewidth, trim=0cm 0cm 0cm 0cm]{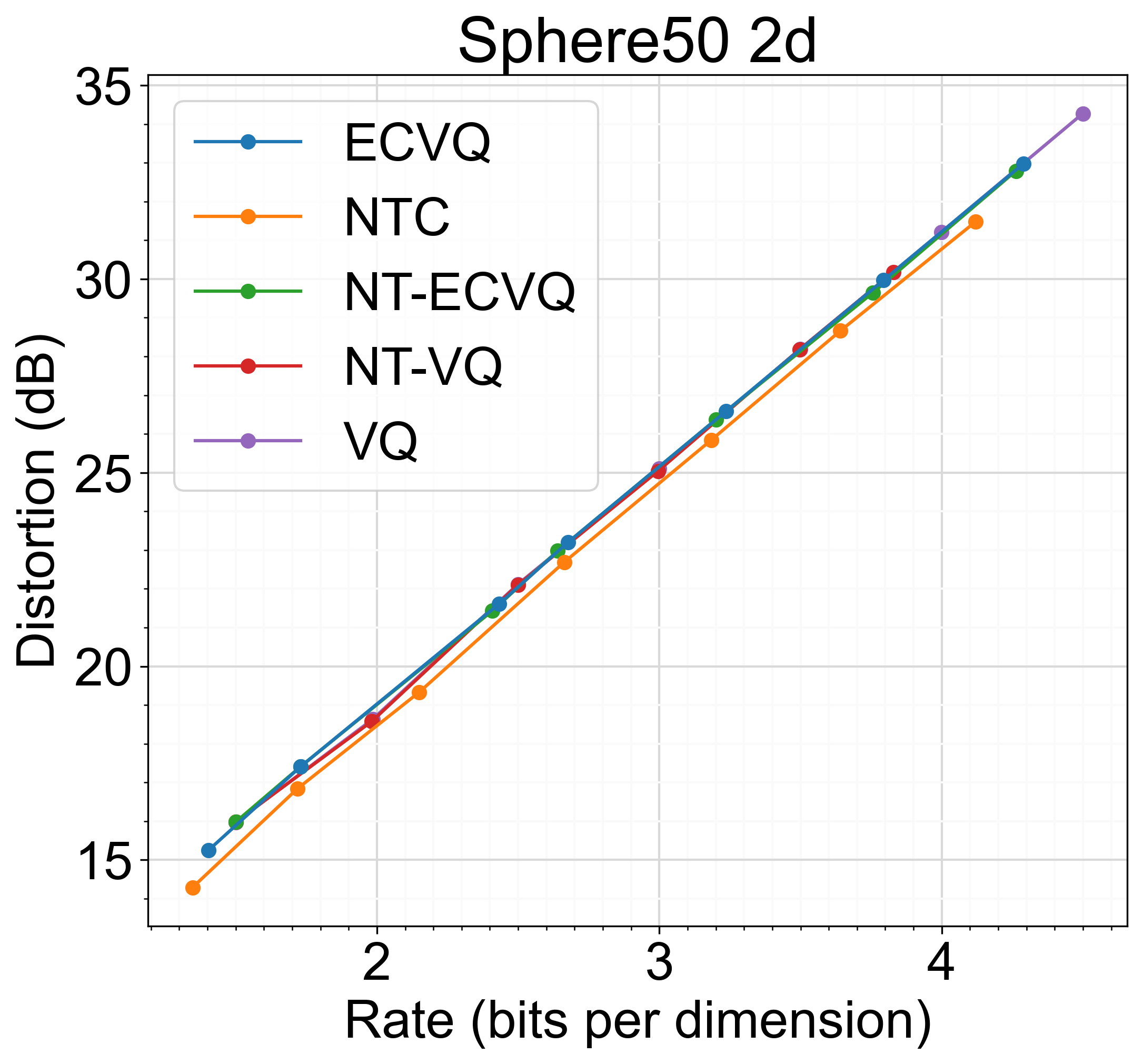}
\includegraphics[width=0.33\linewidth, trim=0cm 0cm 0cm 0cm]{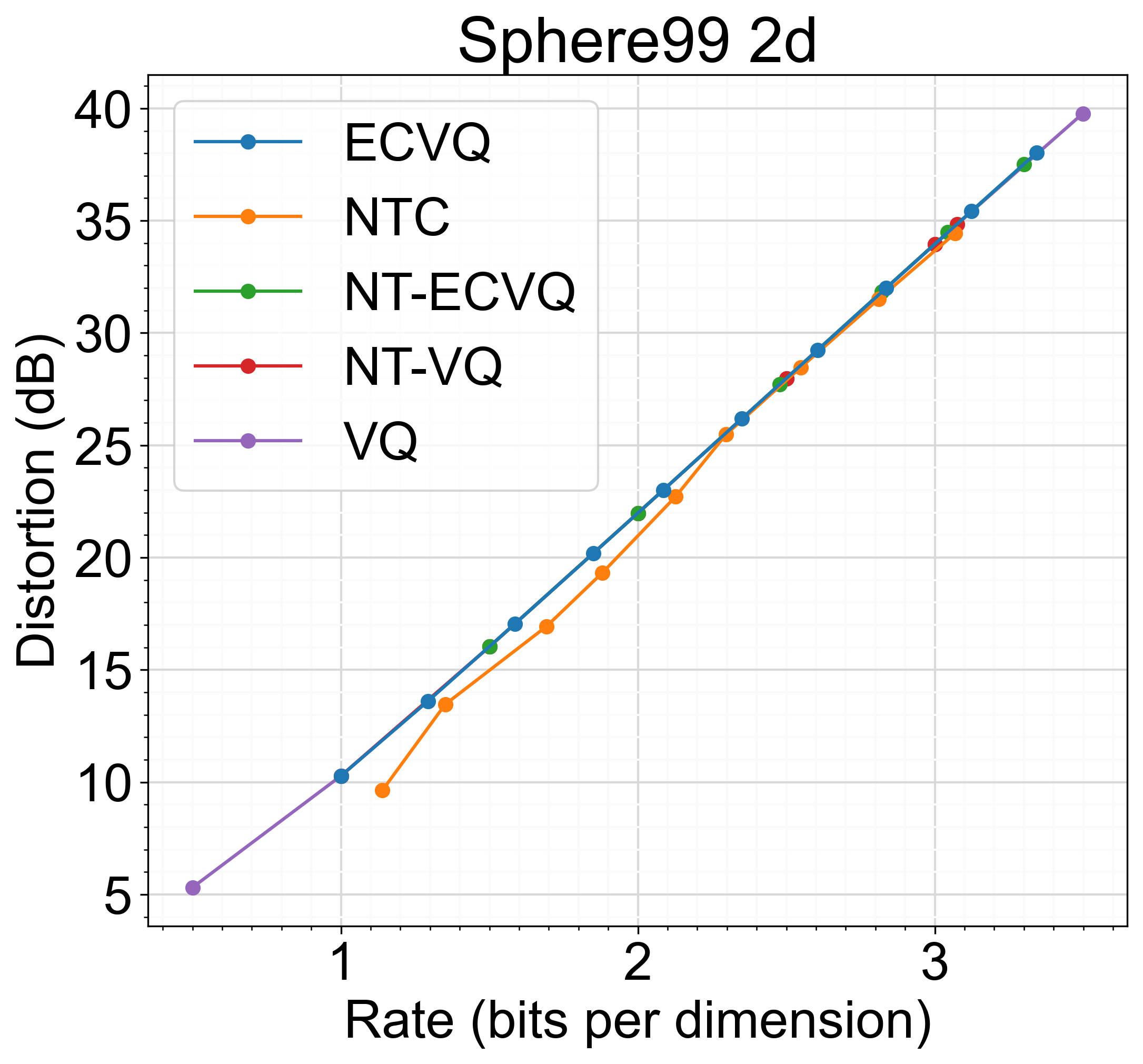}
\end{subfigure}
\caption{RD curves on toy sources.}
\label{fig:rd_toy_sources}
\end{figure*}

\begin{figure*}[th]
 \centering
  \begin{subfigure}{0.8\linewidth}
 \includegraphics[width=\linewidth, trim=0cm 0cm 0cm 0cm]{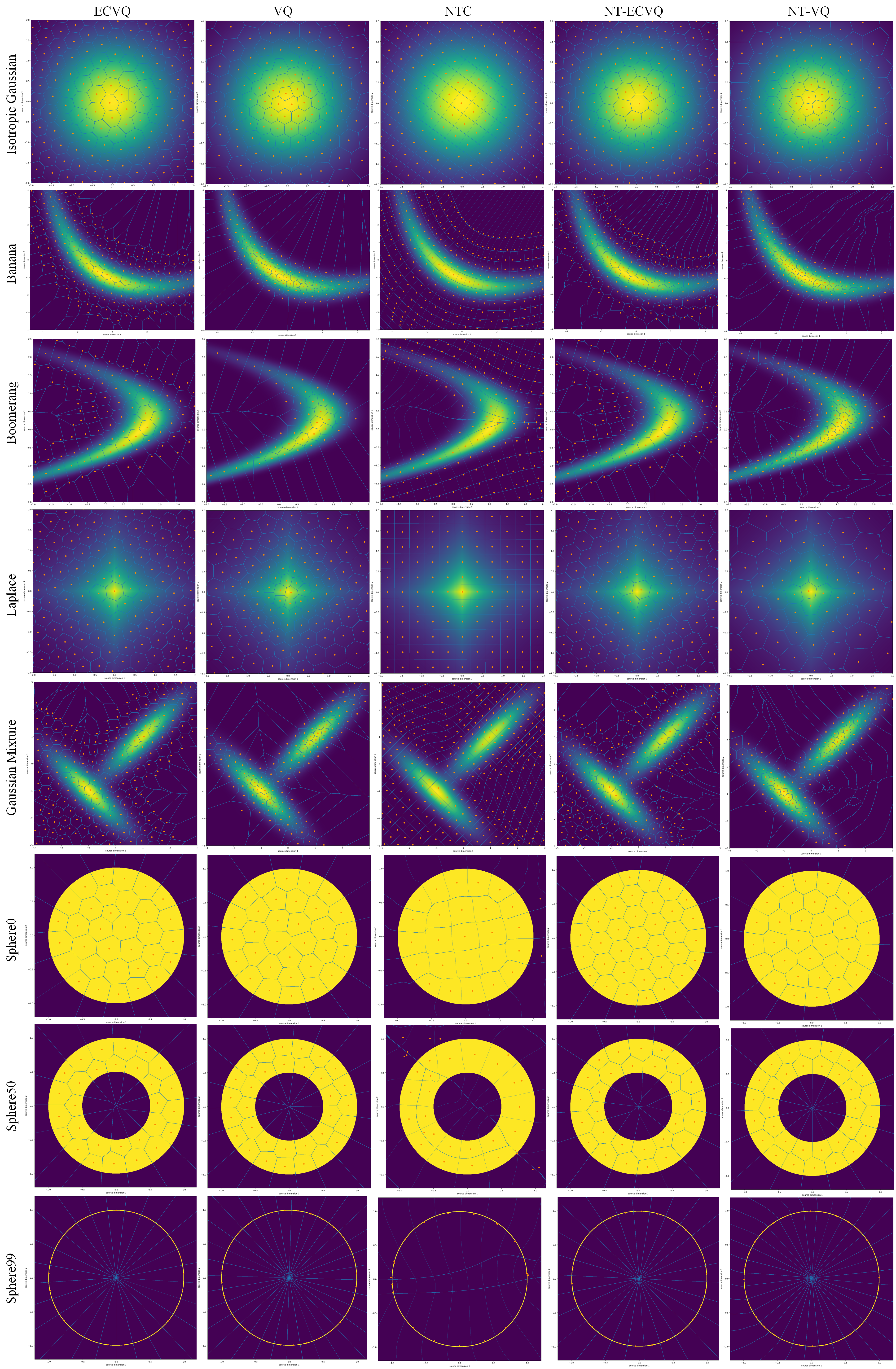}
 \end{subfigure}
\caption{Visualization of quantization results.}
\label{fig:vis_quant_plus}
\end{figure*}

In Figure~\ref{fig:rd_toy_sources} we provide the RD curves.
We show the visualization of quantization results in Figure~\ref{fig:vis_quant_plus}.

\paragraph{Space-filling advantage.}
As mentioned in the main paper, the space-filling advantage increases with dimension (see the RD results of 2-d, 4-d, 8-d and 16-d isotropic Gaussian distributions in Figure~\ref{fig:rd_toy_sources}).

\paragraph{Memory advantage.}
The benefits of memory advantage are significant in the distributions of Boomerang, Gaussian Mixture, Sphere0, Sphere50, and Sphere99. 
All these distributions have highly nonlinear correlations. 

\paragraph{ECVQ vs. VQ.}
ECVQ is better than VQ on all the distributions. 
An observation is that the gap between VQ and ECVQ is very small on the sphere-like uniform distributions (Sphere0, Sphere50, Sphere99).
It is probably because the probabilities of quantization centers on uniform distribution are similar to each other, which decreases the impact of rate bias in ECVQ encoding.
Another observation in Figure~\ref{fig:vis_quant_plus} is that the quantization cells/regions of VQ in high probability area are much smaller than that in low probability area.
In contrast, in ECVQ, the quantization regions in different density area have similar sizes. The reason is that the quantization boundaries in ECVQ shift from high probability region to low probability region, enlarging the size of high-probability region.

\paragraph{ECVQ vs. NT-ECVQ.}

In Figure~\ref{fig:vis_quant_plus}, a major difference between the ECVQ and NT-ECVQ is that NT-ECVQ warps the source space by nonlinear transform, making the quantization boundaries into curves. 
Moreover, as shown in Figure~\ref{fig:rd_toy_sources}, NT-ECVQ optimized with Equation~\ref{eq:loss-ntecvq} has a comparable RD performance to ECVQ, demonstrating the effectiveness of the proposed training loss.

\paragraph{NT-ECVQ vs. NT-VQ.} 
We have verified the effectiveness of entropy-constrained quantization in source space. 
What about the effectiveness in latent space?
Can nonlinear transform learn to approximate the shift of quantization boundaries in ECVQ?
By comparing the performance of NT-ECVQ and NT-VQ, we show that ECVQ is also important in latent space.
Despite nonlinear transform approximate ECVQ well on 1-d distributions~\cite{NTC}, it frequently fails on 2-d distributions at most of the rate points.
As shown in Figure~\ref{fig:vis_quant_plus}, the shift of quantization boundaries is not observed in the quantization results of NT-VQ.

\section*{B. On Neural Images }

\subsection*{B.1. Baseline}

\paragraph{BPG}
The RD results of BPG is obtained from CompressAI~\cite{CompressAI} by running the following command:

\begin{equation}
\begin{aligned}
&\rm python\ \texttt{-}m \ compressai.utils.bench\ bpg\ [dataset] \\
&\rm \texttt{-}q\ 15\ 17\ 19\ 21\ 23\ 25\ 27\ 29\ 31\ 33\ 35\ 37\ 39\ 41\ 43\ 45
\notag  
\end{aligned}
\end{equation}

\paragraph{VTM-18.2} 
The software VTM-18.2 is downloaded from \hyperlink{https://vcgit.hhi.fraunhofer.de/jvet/VVCSoftware_VTM}{https://vcgit.hhi.fraunhofer.de/jvet/VVCSoftware\_VTM}. We first convert the RGB images into YUV444:
\begin{equation}
\begin{aligned}
&\rm ffmpeg \ \texttt{-}i \ [IMGfile] \ \texttt{-}{pix\_fmt} \ yuv444p \ [YUVfile]
\notag       
\end{aligned}
\end{equation}
We then encode the YUV file into bitstream:
\begin{equation}
\begin{aligned}
&\rm EncoderAppStatic \ \texttt{-}{i} \ [YUVfile] \ \texttt{-}{c} \ [CFGfile] \ \texttt{-}{q} \ [QP] \\ 
&\rm \texttt{-}{o}  \ [OUTfile] \ \texttt{-}{b} \ [BINfile] \ \texttt{-}{wdt} \ [W] \ \texttt{-}{hgt} \ [H] \ \texttt{-}{fr} \ 1 \ \texttt{-}{f} \ 1 \\ 
&\rm \texttt{--}{InputChromaFormat=444} \ \texttt{--}{InputBitDepth=8} \\ 
&\rm \texttt{--}{ConformanceWindowMode=1}
\notag       
\end{aligned}
\end{equation}
And the decoding command is:
\begin{equation}
\begin{aligned}
&\rm DecoderAppStatic \ \texttt{-}{o}  \ [OUTfile] \ \texttt{-}{b} \ [BINfile] \ \texttt{-}{d} \ 8
\notag
\end{aligned}
\end{equation}
Finally, we convert the reconstructed YUVfile into RGB images for evaluation:
\begin{equation}
\begin{aligned}
&\rm ffmpeg  \ \texttt{-}{s} \ [W]x[H] \ \texttt{-}{pix\_fmt} \ yuv444p \ \texttt{-}{i} \ [OUTfile] \\
&\rm [IMGfile]
\notag       
\end{aligned}
\end{equation}

\paragraph{Learning-based methods}
The results of Cheng-CVPR20~\cite{Cheng-CVPR2020}, Minnen-NIPS2018~\cite{Minnen-NIPS2018} and Balle-ICLR18~\cite{Balle-ICLR2018} are evaluated based on the reimplementation from CompressAI~\cite{CompressAI}.
The complexity results of Zou-CVPR22~\cite{Zou-CVPR2022} and Zhu-CVPR22~\cite{Zhu-CVPR2022} are evaluated based on their official implementation.
The other results are obtained from the paper or the authors.

\subsection*{B.2. Architecture}
\begin{figure*}[th]
 \centering
  \begin{subfigure}{\linewidth}
 \includegraphics[width=\linewidth, trim=0cm 0cm 0cm 0cm]{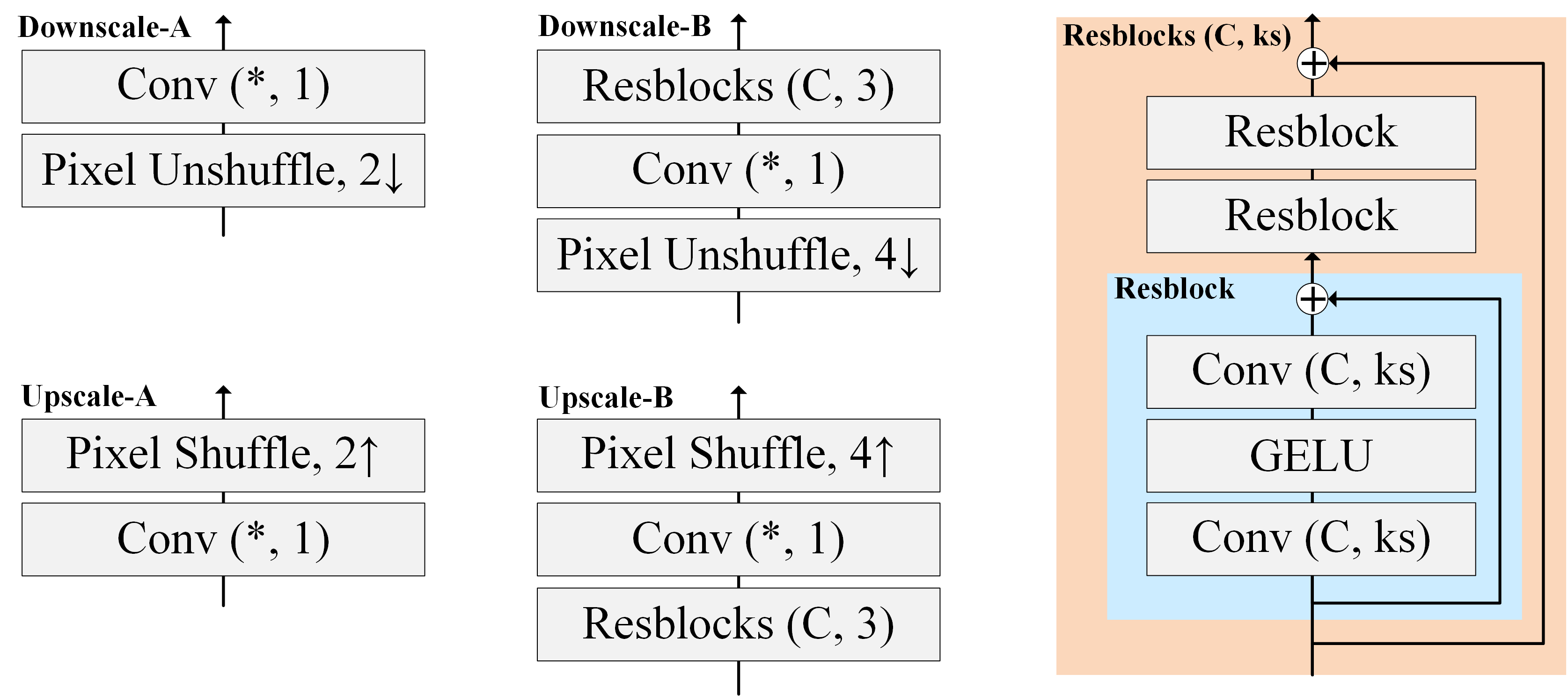}
 \end{subfigure}
\caption{Architecture of Downscale and Upscale layers. ``Conv (C, ks)''  refers to a convolutional layer with C output channels and a kernel size of ks. ``Conv (*, 1)'' refers to a $1\times 1$ convolutional layer with ``*'' output channels which adaptively changes for requirements.  Pixel Unshuffle is a space-to-depth layer (for downsampling) and Pixel Shuffle is a depth-to-space layer (for upsampling)~\cite{Pixelshuffle}. (Right) ``Resblocks (C, ks)'' refers to the Resblock layer consisting of multiple ``Conv (C, ks)''. $C$ is set to 192 for all rate points.}
\label{fig:arch_downupscale}
\end{figure*}

\paragraph{Downscale \& upscale layers.}
The detailed architecture of Downscale and Upscale layers are shown in Figure~\ref{fig:arch_downupscale}.
We use Pixel Shuffle~\cite{Pixelshuffle} instead of strided convolution for upsampling.
Downscale-B is enhanced with Resblocks for learning nonlinear vector representation.

\begin{figure*}[th]
 \centering
  \begin{subfigure}{\linewidth}
 \includegraphics[width=\linewidth, trim=0cm 0cm 0cm 0cm]{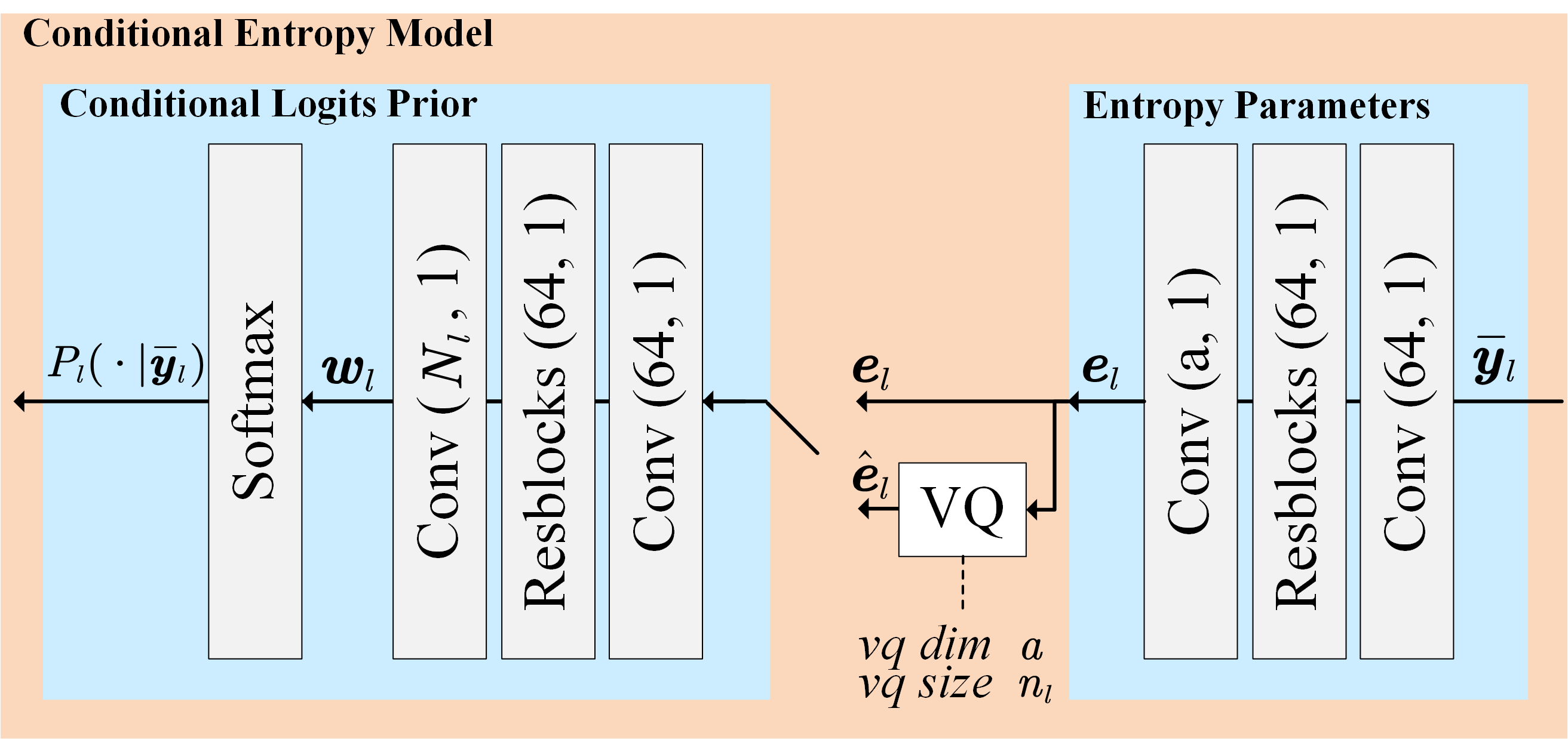}
 \end{subfigure}
\caption{Architecture of Conditional Entropy Model (CEM). }
\label{fig:arch_cem}
\end{figure*}

\paragraph{Conditional entropy model (CEM).}
The architecture of CEM at layer $l$ is shown in Figure~\ref{fig:arch_cem}.
CEM consists of three parts: 
1) an entropy parameter module that generates the prior parameters $\boldsymbol{e}_l$, 
2) a vector quantizer that maps $\boldsymbol{e}_l$ into a finite set, 
and 3) conditional logit prior that outputs discrete probability distribution given the prior parameters $\boldsymbol{e}_l$ or $\boldsymbol{\hat{e}}_l$.
The use of VQ in CEM ensures that all possible probability distributions in a layer can be indexed by a distribution table. 
This distribution table is known to both encoder and decoder after training.
Instead of generating probability distributions dynamically, we simply lookup the distribution table with the VQ index for a much faster entropy coding process (about 6x faster).
Currently, the VQ in CEM brings about 0.15dB drop, which can be further optimized in the future.
In practice, we first train the model without the VQ in CEM, and then finetune the full model.

\paragraph{Model configurations}
We provide the detailed model configurations in Table~\ref{tab:model-cfg}.
For layer $l$, $H_l \times W_l \times C_l$ is the feature size.
$b_l$ is the block size used in the VT units and product VQ.
$N_l$ and $k_l$ are the VQ codebook size and VQ dimension, respectively.
$T_l$ is the iteration milestones for progressive initialization.
Besides, $\delta T$ is set to 10k.

\begin{table}[h]
\begin{center}
\begin{tabular}{|c|c|c|c|c|c|c|c|}
\hline 
Layer & $H_l$ & $W_l$ & $C_l$ & $b_l$ & $N_l$ & $k_l$ & $T_l$ \\ \hline
1  & \multirow{6}{*}{$\frac{H}{16}$} & \multirow{6}{*}{$\frac{W}{16}$} & \multirow{16}{*}{192} & \multirow{16}{*}{4} & \multirow{6}{*}{512} & \multirow{6}{*}{16} & \multirow{6}{*}{0.0M}  \\ 
\cline{1-1} 
2   &  &  &  &  &  &  & \\ 
\cline{1-1} 
3   &  &  &  &  &  &  & \\ 
\cline{1-1} 
4   &  &  &  &  &  &  & \\ 
\cline{1-1} 
5   &  &  &  &  &  &  & \\ 
\cline{1-1} 
6   &  &  &  &  &  &  & \\ 
\cline{1-3} \cline{6-8}
7   & \multirow{6}{*}{$\frac{H}{8}$} & \multirow{6}{*}{$\frac{W}{8}$} &  &  & \multirow{6}{*}{256} & \multirow{6}{*}{8} & \multirow{6}{*}{0.2M}\\
\cline{1-1} 
8   &  &  &  &  &  &  & \\ 
\cline{1-1} 
9   &  &  &  &  &  &  & \\ 
\cline{1-1} 
10  &  &  &  &  &  &  & \\ 
\cline{1-1} 
11  &  &  &  &  &  &  & \\ 
\cline{1-1} 
12  &  &  &  &  &  &  & \\ 
\cline{1-3} \cline{6-8}
13  & \multirow{4}{*}{$\frac{H}{4}$} & \multirow{4}{*}{$\frac{W}{4}$} &  &  & \multirow{4}{*}{128/256} & \multirow{4}{*}{4} & \multirow{4}{*}{0.4M}\\
\cline{1-1} 
14  &  &  &  &  &  &  & \\ 
\cline{1-1} 
15  &  &  &  &  &  &  & \\ 
\cline{1-1} 
16  &  &  &  &  &  &  & \\ 
\hline
\end{tabular}
\end{center}
\caption{Detailed configurations of different quantization layers. }
\label{tab:model-cfg}
\end{table}

\subsection*{B.3. Experiments}
\begin{figure}[th]
\centering
\begin{subfigure}{\linewidth}
\includegraphics[width=\linewidth, trim=0cm 0cm 0cm 0cm]{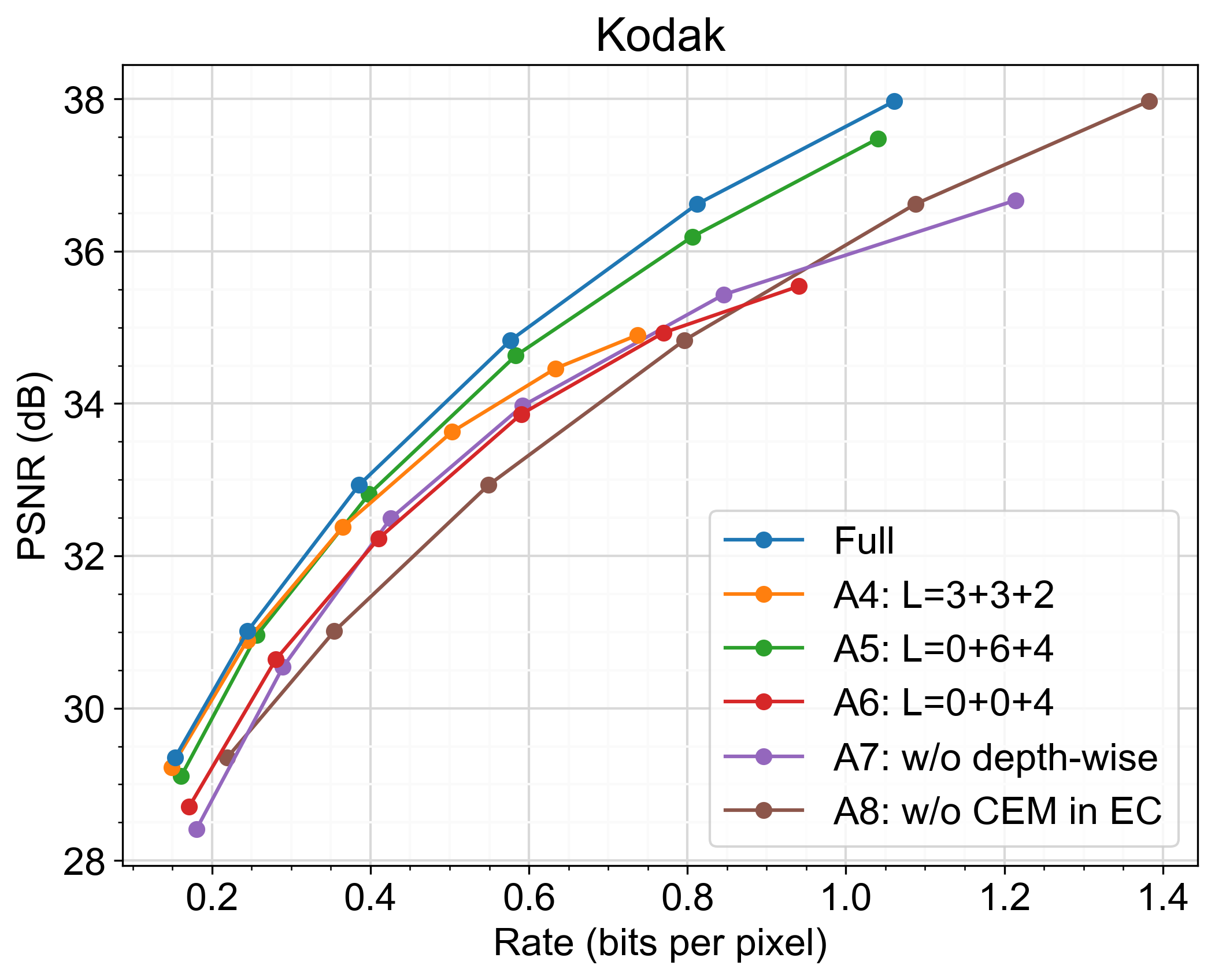}
\end{subfigure}
\caption{Ablation study on Kodak dataset.}
\label{fig:rd_ablation}
\end{figure}

In this section, we provide additional ablation studies, model properties and visualization of reconstruction.
For ablation studies, we provide the BD-rate in Table~\ref{tab:ablation-plus} and the RD curves in Figure~\ref{fig:rd_ablation}.
BD-rate is calculated using the software from \hyperlink{https://github.com/Anserw/Bjontegaard_metric}{https://github.com/Anserw/Bjontegaard\_metric}.
All models are optimized with 128 $\times$ 128 image patches.
\begin{table}[h]
\begin{center}
\begin{tabular}{lcc}
\hline 

             & BD-rate & Parameters (M) \\ \hline
Full model   & 0.0\% & 12.8 \\
A4: L=3+3+2  & 7.7\%  & 8.7  \\ 
A5: L=0+6+4  & 6.8\%  & 8.9  \\ 
A6: L=0+0+4  & 27.1\% & 5.8  \\ 
A7: w/o depth-wise  &  29.4\%  & 9.5  \\ 
A8: w/o CEM in EC  &  39.6\%  & 12.8  \\ 
\hline
\end{tabular}
\end{center}
\caption{Ablation study on Kodak. BD-rate is computed over the full models (smaller is better). }
\label{tab:ablation-plus}
\end{table}

\paragraph{The number of quantization layers.}
As mentioned in the main paper, the number of quantization layers influence the rate-distortion-complexity trade-off.
The full model consists of 16 quantization layers with $(L_1, L_2, L_3) = (6, 6, 4)$.
We change $(L_1, L_2, L_3)$ to $(3, 3, 2)$, $(0, 6, 4)$ and $(0, 0, 4)$ and train 3 model variants, which is named ``A4: L=3+3+2'', ``A5: L=0+6+4'' and ``A6: L=0+0+4'', respectively.
It can be observed that
1) the models with fewer quantization layers perform worse at high rate points, 
and 2) the low-resolution quantization layers (numbered by $L_1$ or $L_2$) which capture global correlation have a significant impact at all rate points.

\paragraph{Depth-wise Block FC vs. Block FC.}
We investigate the impact of the proposed Depth-wise Block FC layer.
We build a variant named Block FC where all channels use the same transformation matrix.
Block FC is similar to Block-DCT with a learnable transformation matrix.
The performance of Block FC (noted as ``A7: w/o depth-wise'') is much worse than the full model.

\paragraph{CEM in entropy coding.}
CEM plays an important role in removing inter-vector redundancy across different quantization layers.
We investigate rate savings when disabling CEM in entropy coding.
Instead of retraining the whole model, we simply replace the CEM in a trained model with the unconditional entropy model (UEM) shown in Equation~\ref{eq:UEM}.
Then UEM is optimized using the rate loss only for entropy coding.
The results (``A8: w/o CEM in EC'') demonstrate the significant rate savings of CEM, which is not considered in previous works~\cite{Zhu-CVPR2022,Agustsson-NIPS2017} with VQ.

\begin{figure*}[th]
 \centering
  \begin{subfigure}{0.8\linewidth}
 \includegraphics[width=\linewidth, trim=0cm 0cm 0cm 0cm]{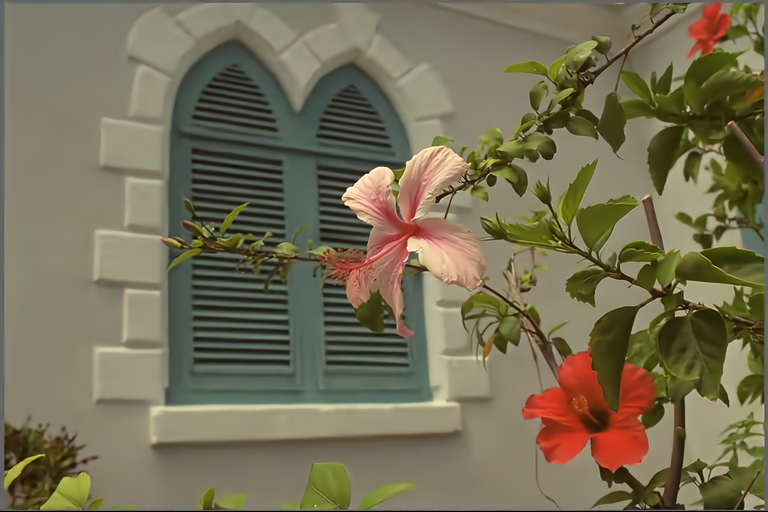}
 \caption{Ours. Bpp=0.2626, PSNR=35.23dB.}
 \end{subfigure}
   \begin{subfigure}{0.8\linewidth}
 \includegraphics[width=\linewidth, trim=0cm 0cm 0cm 0cm]{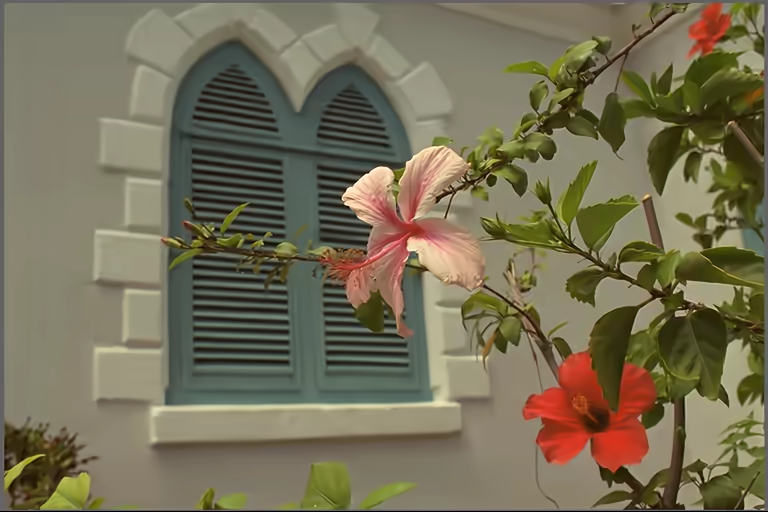}
 \caption{VVC. Bpp=0.2761, PSNR=35.32dB.}
 \end{subfigure}
\caption{Comparison between the proposed method and VVC on ``Kodim07.png''.}
\label{fig:vis_sub1}
\end{figure*}

\begin{figure*}[th]
 \centering
  \begin{subfigure}{0.8\linewidth}
 \includegraphics[width=\linewidth, trim=0cm 0cm 0cm 0cm]{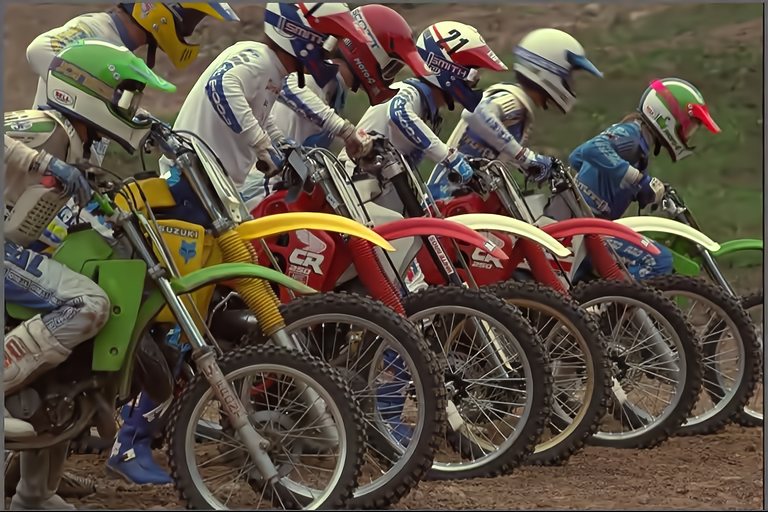}
 \caption{Ours. Bpp=0.6574, PSNR=31.45dB.}
 \end{subfigure}
   \begin{subfigure}{0.8\linewidth}
 \includegraphics[width=\linewidth, trim=0cm 0cm 0cm 0cm]{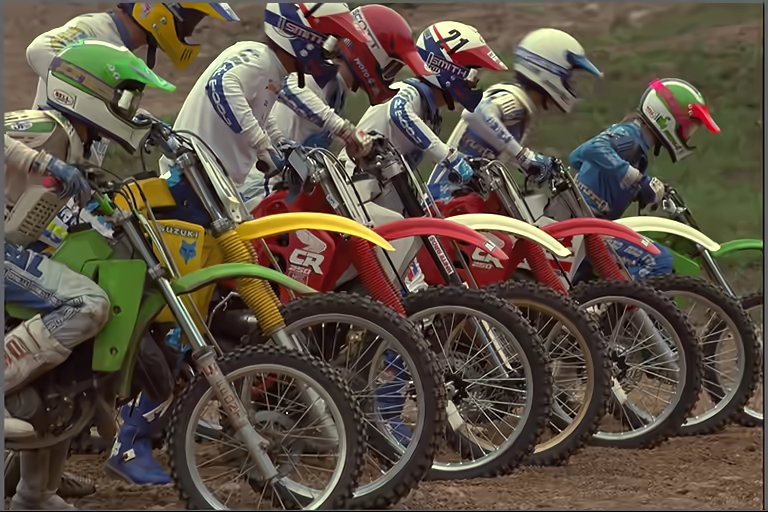}
 \caption{VVC. Bpp=0.7051, PSNR=31.60dB.}
 \end{subfigure}
\caption{Comparison between the proposed method and VVC on ``Kodim05.png''.}
\label{fig:vis_sub2}
\end{figure*}

\paragraph{Subjective comparison.}
In Figure~\ref{fig:vis_sub1} and Figure~\ref{fig:vis_sub2}, we provide the subjective comparison between our method and VVC. 
Our reconstruction has a slightly better subjective quality with a smaller bpp.

\end{document}